\documentclass[final,5p,times,twocolumn,authoryear,round]{elsarticle}

\usepackage[authoryear, round]{natbib}
\usepackage{graphicx}
\usepackage{amssymb}
\usepackage{multirow}
\usepackage{amsmath}
\usepackage{xcolor}
\usepackage[table]{xcolor}
\usepackage{tabularx}
\newcolumntype{Y}{>{\centering\arraybackslash}X} 
\usepackage{algorithm} 

\usepackage{algpseudocode} 

\usepackage{amsmath}

\journal{ESWA}

\begin{document}

\begin{frontmatter}

\title{CAWM-Mamba:A Unified Model for Infrared-Visible Image Fusion and Compound Adverse Weather Restoration}

\author[lable1]{Huichun Liu}
\ead{2112355008@stu.fosu.edu.cn}

\author[lable1,lable2]{Xiaosong Li\corref{cor}}
\ead{lixiaosong@buaa.edu.cn}

\author[lable1]{Zhuangfan Huang}
\ead{2112455033@stu.fosu.edu.cn}

\author[lable3]{Tao Ye}
\ead{yetao@cumtb.edu.cn}

\author[lable1]{Yang Liu}
\ead{ly25@fosu.edu.cn}

\author[lable1]{Haishu Tan }
\ead{tanhaishu@fosu.edu.cn}

\cortext[cor]{Corresponding author}

\address[lable1]{ School of Physics and Optoelectronic Engineering, Foshan University, Foshan 528225, China}

\address[lable2]{Guangdong-HongKong-Macao Joint Laboratory for Intelligent Micro-Nano Optoelectronic Technology, Foshan 528225, China}

\address[lable3]{School of Mechanical Electronic and Information Engineering, China University of Mining and Technology, Beijing 100083, China}

\begin{abstract}
  Multimodal Image Fusion (MMIF) integrates complementary information from various modalities to produce clearer and more informative fused images. MMIF under adverse weather is particularly crucial in autonomous driving and UAV monitoring applications. However, existing adverse weather fusion methods generally only tackle single types of degradation such as haze, rain, or snow, and fail when multiple degradations coexist (e.g., haze\&rain, rain\&snow). To address this challenge, we propose Compound Adverse Weather Mamba (CAWM-Mamba), the first end-to-end framework that jointly performs image fusion and compound weather restoration with unified sharedweights.
  Our network contains three key components: (1) a Weather-Aware Preprocess Module (WAPM) to enhances degraded visible features and extracts global weather embeddings; (2) a Cross-modal Feature Interaction Module (CFIM) to facilitate the alignment heterogeneous modalities and exchange of complementary features across modalities; and (3) a Wavelet Space State Block (WSSB) that leverages wavelet-domain decomposition to decouple multi-frequency degradations. 
  WSSB includes Freq-SSM, a module that models anisotropic high-frequency degradation without redundancy, and a unified degradation representation mechanism to further improve generalization across complex compound weather conditions.
  Extensive experiments on the AWMM-100K benchmark and three standard fusion datasets demonstrate that CAWM-Mamba consistently outperforms state-of-the-art methods in both compound and single-weather scenarios. In addition, our fusion results excel in downstream tasks covering semantic segmentation and object detection, confirming the practical value in real-world adverse weather perception. The source code will be available at https://github.com/Feecuin/CAWM-Mamba.

\end{abstract}

\end{frontmatter}
\section{Introduction}
The rapid development of optical imaging and multimodal information processing has made image fusion a critical research frontier. Single-sensor images often fail to capture all the information from a scene. Infrared (IR) images, for instance, can penetrate smoke and obstructions, highlighting thermal targets, but their low resolution and lack of texture details limit their effectiveness in certain applications. Conversely, visible images (VI) provide high-resolution texture and color details, but they are highly susceptible to light attenuation and adverse weather conditions, leading to information loss or distortion. Therefore, Infrared and Visible Image Fusion (IVIF)\citep{325-tip,325-my,IVIF3,IVIF7} has emerged as a core technology in multimodal information processing. IVIF integrates the complementary information from IR and VI spectral characteristics to generate fused images with higher information density, better visual perception, and stronger environmental adaptability. This collaborative fusion not only overcomes the limitations of single-modal images but also significantly enhances target detection, scene understanding, and decision-making capabilities, making it widely applicable in fields such as military reconnaissance, autonomous driving, medical diagnosis, and environmental monitoring.

However, most current IVIF methods are developed under ideal conditions and primarily focus on interference-free scenarios. As practical applications continue to advance, such as in autonomous driving and drone monitoring, adverse weather conditions (such as haze, rain, and snow) have significantly affected the quality of infrared and visible light images, thereby impacting the fusion results. Therefore, although existing methods perform well under ideal conditions, they often fail to provide reliable fusion outcomes in complex weather scenarios.  

Under adverse weather conditions, the imaging characteristics and fusion effects of infrared and visible light images are significantly impacted. Visible images are highly vulnerable to adverse weather (e.g., haze, rain, snow), often suffering from detail loss or distortion. Infrared imaging, though more resilient due to thermal radiation, lacks fine textures and is prone to noise, resulting in limited resolution. With advancements in research, some methods\citep{AWFusion,Text-IF,TG} have begun to address image fusion under single weather degradation conditions, such as fusion recovery for interference caused by single environmental factors like rain, haze, or snow. Although some progress has been made, these methods still struggle to effectively handle the combined effects of multiple degradation factors. Particularly under compound weather interferences, such as rain and haze or snow and rain, existing methods often fail to maintain optimal fusion results, leading to a decline in fusion quality.

Current fusion methods primarily focus on single-degradation scenarios and are ill-equipped to handle the complex interference caused by compound weather conditions. Furthermore, many existing approaches rely on manual preprocessing, which significantly limits their inference efficiency and practical applicability. Therefore, developing a unified model capable of jointly handling compound degradation and multimodal fusion is of great practical importance. To address these challenges, we propose the Compound Adverse Weather Mamba (CAWM-Mamba), an innovative end-to-end network designed to perform both image fusion and restoration under compound weather conditions. Unlike previous approaches limited to single-weather restoration, CAWM-Mamba unifies multiple degradations within a single framework, ensuring robust fusion even in complex environments.

1) We preprocess the degraded visible light images, enhance the features of the degraded visible light images, and extract global weather information as the embedding to improve the quality of image fusion. 2) We conduct cross-modal interaction between infrared and visible light images. We fuse the complementary characteristics of infrared and visible light, and utilize multi-scale pooling and feature interaction to enhance the precise complementation and alignment of structure and texture. 3) The wavelet-domain decomposition technique is used to decouple compound degradation, and by designing the Freq-SSM based on high-frequency components' directional features, the model's ability to handle multi-directional degradation interference under complex weather conditions is further enhanced. At the same time, the unified degradation representation is learned, improving the model's generalization capability. \textbf{Our contributions are summarized as follows:}
\begin{itemize}
    \item We propose CAWM-Mamba, an innovative end-to-end framework for image fusion and restoration under compound weather conditions, capable of handling multiple degradations simultaneously to ensure robust fusion in complex environments.
\end{itemize}

\begin{itemize}
    \item We introduce the Weather-Aware Preprocess Module (WAPM) and the Cross-Modal Feature Interaction Module (CFIM), which together enhance the fusion process by embedding global weather information, improving visible light feature quality, and effectively aligning features from infrared and visible images through multi-scale pooling and feature interaction.
\end{itemize}

\begin{itemize}
    \item We develop the Wavelet Space State Block (WSSB), a novel technique for decoupling multi-frequency degradation patterns using wavelet-domain decomposition, improving the model’s ability to handle complex degradation and enhancing its generalization across various weather conditions.
\end{itemize}

\begin{itemize}
    \item Extensive experiments on AWMM-100K and three standard datasets show CAWM-Mamba surpasses SOTA methods under both single and compound weather. It also boosts downstream segmentation and detection, proving its real-world value.
\end{itemize}

The remainder of this paper is organized as follows. Section 2 gives a review of the related work. Section 3 details the proposed CAWM-Mamba. Section 4 presents the experimental validations. Section 5 concludes this work and givesitations and future prospects.

\section{Related Work}
\subsection{Clean Image Fusion Methods.}
Existing image fusion methods can be broadly categorized into traditional  based and deep learning-based approaches. The former typically designed specific fusion strategies in either the transform domain or the spatial domain\citep{xiaosong1,xiaosong2}. Among transform-domain methods, multiscale transforms\citep{duochidu1-CM-MCNet,duochidu2-MRF-Net} are widely used. These rely on predefined multiscale decomposition tools to simulate the coarse-to-fine perceptual characteristics of the human visual system. For instance, pyramid decomposition and wavelet transform enable multi-resolution representations of images. In contrast, spatial-domain methods perform fusion directly at the pixel, block, or region level by weighting, selecting, or combining pixel values or local features from the source images to generate fused results. In addition, sparse representation is another important traditional approach. It assumes that image patches can be sparsely encoded under an over-complete dictionary, thereby extracting more discriminative features, and several fusion methods\citep{xiaosong1,xiaosong2} based on sparse representation have also been proposed. However, transform-domain methods\citep{jin1-WANG2024102414,jin2-LEITE2025103339} often cause information loss during decomposition and reconstruction; spatial-domain methods\citep{kong4,kong5} are sensitive to block size and neighborhood dimensions; and sparse representation methods are computationally expensive. Moreover, these handcrafted strategies show limited adaptability under complex environmental interference, making it difficult to meet higher performance requirements.

In contrast, deep learning-based methods\citep{325-yixue,325-yixue1,IVIF6,kungong-pami,kungong3,ESWA-jie,ESWA-ronghe,ESWA-ronghe1,325-xu} can automatically learn feature representations through architectures such as Convolutional Neural Networks (CNNs), Generative Adversarial Networks (GANs), and Transformers, demonstrating significant advantages in multimodal tasks such as infrared–visible fusion and medical imaging. CNN-based methods extract features via convolution operations and apply fusion rules to emphasize feature preservation and detail enhancement, such as DenseFuse\citep{DenseFuse}, U2Fusion\citep{U2Fusion}, and LRRNet\citep{LRRNet}. GAN-based methods employ adversarial training to balance modality-specific features, such as FusionGAN\citep{FusionGAN} and DDcGAN\citep{DDcGAN}. Transformer-based methods focus on cross-modal feature decomposition, semantic modeling, and long-range dependency learning, exemplified by CDDFuse\citep{CDDFuse}. Recently, improvements have incorporated semantic priors, low-light enhancement, or task-guided mechanisms (e.g., DIVFusion\citep{DIVFusion}, FS-Diff\citep{FS-Diff}, TIM\citep{TIM}), along with fusion models tailored for downstream perception tasks (e.g., TarDal\citep{TarDal}, SegMiF\citep{SegMiF}, UMFusion\citep{UMFusion}). These methods have achieved promising results across diverse applications and are gradually driving fusion models toward greater generalization and robustness.

However, the above competing methods assume that the input images are clean, overlooking the common problem of weather-induced degradation in real-world scenarios. As a result, their applicability in complex environments remains limited.

\subsection{Single Degradation Image Restoration.}
Significant progress has been made in recovering clean images from specific weather degradations. Early works primarily relied on Convolutional Neural Networks (CNNs), but recent advancements have shifted toward Transformer and State Space Model (SSM) architectures to capture long-range dependencies. For instance, Liang et al.\citep{re-4} proposed SwinIR, which integrates the advantages of Swin Transformer for image restoration, while Zamir et al.\citep{re-1} introduced Restormer, an efficient Transformer that computes self-attention across channel dimensions to handle high-resolution images. To address multiple degradation types, Potlapalli et al.\citep{re-2} developed PromptIR, utilizing visual prompts to adaptively guide the network for different weather conditions. More recently, Guo et al.\citep{re-3} proposed MambaIR, establishing a simple yet effective baseline that leverages the linear complexity of Mamba for low-level vision. Additionally, task-specific innovations continue to emerge, such as the channel consistency prior for unsupervised deraining\citep{re-5} and interaction-guided strategies for image dehazing\citep{re-6}. However, these methods typically assume a single degradation type and operate solely on visible spectrum inputs. They often struggle to handle compound weather conditions (e.g., rain streaks coupled with haze) and fail to incorporate complementary thermal information from infrared sensors. This limitation highlights the necessity for a unified framework capable of jointly performing multi-modal image fusion and heterogeneous weather restoration.

\subsection{Image Fusion With Adverse Weather Methods. }
In real-world scenarios, images are often affected by weather-induced degradation factors such as rain, haze and snow. Under such conditions, existing image fusion methods struggle to achieve satisfactory results due to their limited capability in handling degradations. Even high-performance deep learning methods, if they fail to account for image degradation, tend to produce fused images with blurred details and information loss, which stem from the poor quality of the source images. Therefore, integrating image restoration with fusion techniques has become a feasible way to meet the practical demands of complex environments.

In recent years, several studies have made progress in multimodal image fusion and restoration under adverse weather conditions. For example, Text-IF\citep{Text-IF} incorporates semantic text into the fusion task, enabling degradation-aware and interactive fusion guided by semantics. AWFusion\citep{AWFusion} proposes a physics-aware multimodal image fusion model that simulates the degradation process using an atmospheric scattering model and restores clear features via low-rank decomposition.
AWM-Fuse\citep{325-my} adopts a unified architecture that integrates global and local text perception with visual language model technology to realize multimodal image fusion in adverse weather conditions, thus addressing the problem of visual information loss induced by adverse weather and enhancing image quality.

However, these attempts are only designed for a single type of degradation (e.g., handling only rain or haze) and perform poorly in compound degradation scenarios where multiple weather factors coexist. In such cases, different degradation mechanisms often overlap, making the degradation patterns more complex, while existing methods lack a unified modeling capability to address them. Hence, there is an urgent need for a unified framework capable of simultaneously handling multiple degradations and achieving high-quality fusion—this is the focus of our study.
\begin{figure*}
  \centering
   \includegraphics[width=1.0\linewidth]{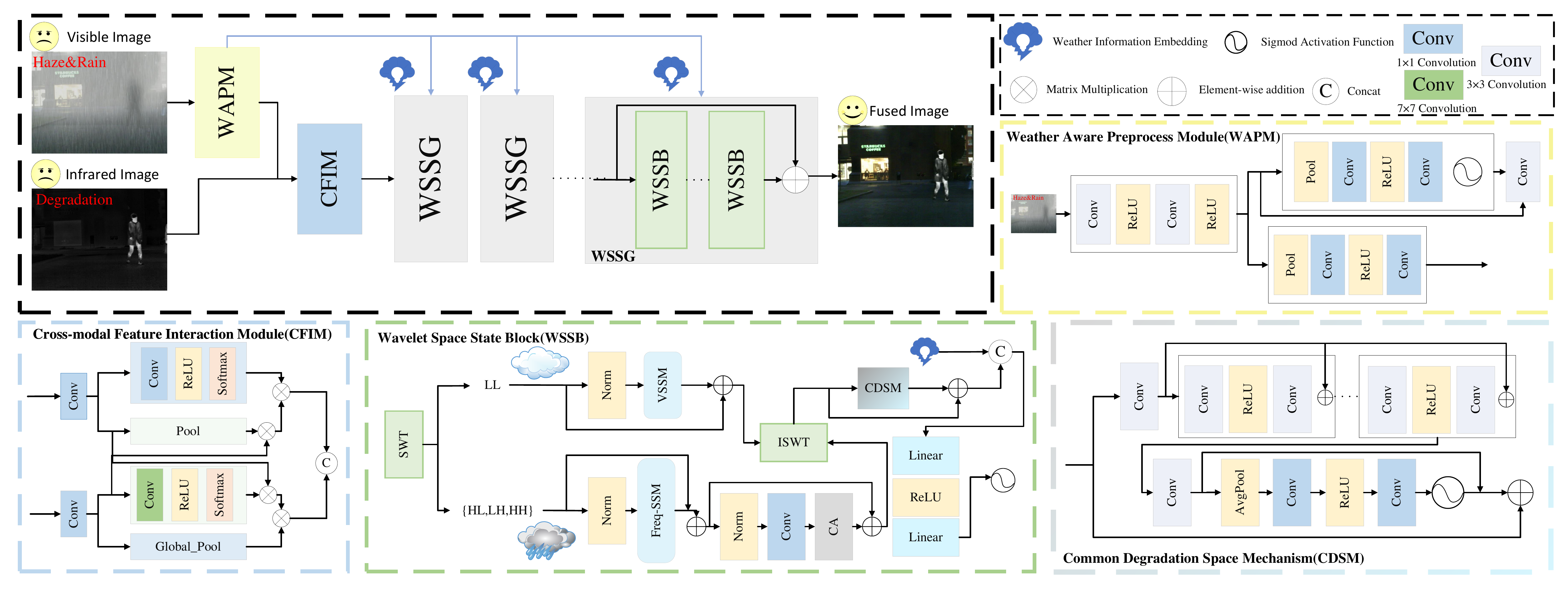}
   \caption{The overall architecture of CAWM-Mamba. It consists of three main components: WAPM, CFIM, and WSSB. Within WSSB, Freq-SSM captures anisotropic high-frequency features, while CDSM learns a unified degradation representation for robust generalization.}
   \label{fig1}
\end{figure*}
\section{Methodology}
We propose CAWM-Mamba, an end-to-end framework for multimodal image fusion and compound weather restoration. The architecture adopts a progressive design (see Figure~\ref{fig1}), where each component is motivated by specific challenges in compound weather scenarios. First, the Weather-Aware Preprocess Module (WAPM) enhances degraded visible features and extracts global weather embeddings. Then, the Cross-modal Feature Interaction Module (CFIM) enables cross-modal alignment and complementary information exchange between modalities. Finally, the Wavelet Space State Block (WSSB) transforms features into the wavelet domain to decouple frequency-specific degradations. Within WSSB, the Freq-SSM captures anisotropic high-frequency details without directional redundancy, while the Common Degradation Space Mechanism (CDSM) learns a unified degradation representation to improve generalization across diverse degradation patterns.

\subsection{Weather Aware Preprocess Module}
In the joint task of image fusion and adverse weather restoration, weather conditions significantly affect the quality of visible light images, thereby influencing the fusion results. Therefore, we propose a Weather Aware Preprocess Module (WAPM), designed to preprocess input weather-interfered visible light images, extract weather-related global feature embeddings, and enhance image features, providing more effective inputs for subsequent fusion and restoration tasks.

The input visible image (VI) first undergoes feature extraction through two convolutional layers and ReLU activation functions to obtain an initial feature map $VI_{\text{feat}}$, which is used to capture weather degradation-related information.
The feature map is compressed into a global feature vector via global average pooling, followed by feature transformation through two convolutional layers and ReLU activation functions to learn inter-channel nonlinear relationships. Finally, channel attention weights are generated by the Sigmoid function for channel-wise weighting of the feature map, achieving enhancement of weather degradation-related features.
\begin{equation}
atten = \text{Sigmoid} \left( W_2 \left( \sigma \left( W_1 \left( \text{GAP}(VI_{\text{feat}}) \right) \right) \right) \right)
\end{equation}
\begin{equation}
VI_{\text{enhance}} = \text{atten} *VI_{\text{feat}}
\end{equation}
where,$W_{\text{i}}$ represents the weight of a 3×3 convolution kernel, and $\sigma$ denotes the ReLU activation function. 

The number of channels of the enhanced feature map $VI_{\text{enhance}}$ is adjusted to 3 to obtain $VI_{\text{out}}$ , which is used for subsequent processing. Meanwhile, to acquire weather-related global information, this part compresses the feature map into a global feature vector through adaptive average pooling, then performs feature transformation via two convolutional layers and ReLU activation functions, and finally outputs a fixed-dimensional global feature embedding $F_{\text{embedding}}\in R^{B \times 48}$ . The global feature embedding contains weather-related global information, helping subsequent modules better understand the overall weather conditions of the image and providing explicit prior guidance for subsequent processing.
\begin{equation}
F_{\text{embedding}} = W_4\bigl(\sigma\bigl(W_3\bigl(\text{GAP}(VI_{\text{enhance}})\bigr)\bigr)\bigr)
\end{equation}

The preprocessing by WAPM not only enhances the features of visible light images but also extracts weather-related global information. The weather feature embedding serves as auxiliary information to help improve the quality of image fusion, providing richer and more effective feature information for subsequent image fusion and adverse weather restoration tasks.
\subsection{Cross-modal Feature Interaction Module}
In image processing under compound weather interference, how to effectively fuse infrared and visible light image features and extract common degradation features is a key issue. To effectively integrate the complementary information of infrared and visible images, we design a Cross-modal Feature Interaction Module(CFIM). The module achieves effective information exchange between the two modalities by means of multi-scale feature extraction and adaptive feature fusion, complementing the advantages of IR and VI images. The module is divided into feature modulation and feature fusion stages.

IR and VI features are processed separately to maintain modality independence. Given the infrared image input and visible light image input , feature extraction and dimension unification are first performed through 1×1 convolution:
\begin{equation}
F_{ir} = \text{Conv1}(I_{ir}),\ F_{vi} = \text{Conv1}(I_{vi})
\end{equation}
where, $F_{\text{ir}},F_{\text{vi}}\in R^{B \times C\times H\times W}$ , where B is the batch size, C is the number of feature channels, and H, W are the height and width of the feature map.

The feature channels of $F_{\text{ir}},F_{\text{vi}}$ are equally divided into two parts, denoted as $F_{\text{ir}}^{mp}, F_{\text{ir}}^{gmp}$ and $F_{\text{vi}}^{ap}, F_{\text{vi}}^{gap}$. $F_{\text{ir}}^{mp}$ uses MaxPooling to obtain $P_{\text{ir}}^{mp}$, which extracts salient features, captures thermal radiation contour information of targets, and preserves features in regions with large temperature gradients. $F_{\text{vi}}^{ap}$ uses AvgPooling to obtain $P_{\text{vi}}^{ap}$ , which extracts texture features, maintains scene detail information, and acquires global brightness and color distributions. Attention weights are then generated through a convolutional network.
\begin{align}
A_{vi}^{ap} &= \text{Softmax} \bigl( \sigma \bigl( \text{Conv7}(P_{ir}^{mp}) \bigr) \bigr), \\
A_{ir}^{mp} &= \text{Softmax} \bigl( \sigma \bigl( \text{Conv7}(P_{vi}^{ap}) \bigr) \bigr)
\end{align}
Meanwhile, $P_{\text{ir}}^{gmp}$ and $P_{\text{vi}}^{gap}$ are obtained by using global max pooling and global average pooling, which captures the global statistical characteristics of the scene to assist cross-modality alignment:
\begin{align}
A_{ir}^{\text{gmp}} &= \text{Softmax}\bigl( \sigma\bigl( \text{Conv1}(P_{ir}^{\text{gmp}}) \bigr) \bigr), \\
A_{vi}^{\text{gmp}} &= \text{Softmax}\bigl( \sigma\bigl( \text{Conv1}(P_{vi}^{\text{gmp}}) \bigr) \bigr)
\end{align}
The saliency of infrared thermal targets enhances the texture response of visible image, while the spatial details of visible light optimize the boundaries of infrared targets, achieving feature-level information exchange. Finally, features carrying different modal information are multiplied by their own channel importance weights.
\begin{equation}
F_{ir}^{out} = \left( A_{ir}^{mp} * F_{ir}^{mp} \right) * A_{ir}^{gmp},\ F_{vi}^{out} = \left( A_{vi}^{ap} * F_{vi}^{mp} \right) * A_{vi}^{gmp}
\end{equation}
Finally, perform dimension concatenation:
\begin{equation}
F^{out} = \text{Concat}[F_{ir}^{out}, F_{vi}^{out}]
\end{equation}

\subsection{Wavelet Space State Block}
The Wavelet Space State Block (WSSB) aims to address the challenges posed by complex weather changes by transforming degradation-containing images into the wavelet domain. Compared to other multiscale representations, the wavelet transform offers distinct advantages for compound degradation modeling. Unlike the Fourier transform, which provides global frequency information but lacks spatial localization, wavelets offer simultaneous time-frequency localization, enabling the model to pinpoint spatially variant degradations like uneven haze or local rain streaks. Furthermore, compared to isotropic methods like Laplacian pyramids, the wavelet transform explicitly decouples features into horizontal, vertical, and diagonal subbands. This directional separation is crucial for handling anisotropic weather artifacts (e.g., driving rain or wind-blown snow), allowing our subsequent Freq-SSM to process degradation along its principal direction. Different types of degradations affect different frequency bands. To effectively separate and process various degradation types, we use wavelet transform for feature decoupling of compound weather, thereby separating the distinct frequency-domain manifestations of weather interferences such as rain, haze, and snow. First, we utilize the Discrete Wavelet Transform (DWT) to decompose the input image $I_{in}$. Specifically, we employ the Haar wavelet with a decomposition level of $J=1$ and apply zero-padding for boundary handling to obtain the low-frequency component $I_{LL}$ and high-frequency components $(I_{HL}, I_{LH}, I_{HH})$: 
\begin{equation}
\{I_{LL}, I_{LH}, I_{HL}, I_{HH}\} = \text{DWT}(I_{\text{in}})
\end{equation}

\begin{figure}
  \centering
   \includegraphics[width=1\linewidth]{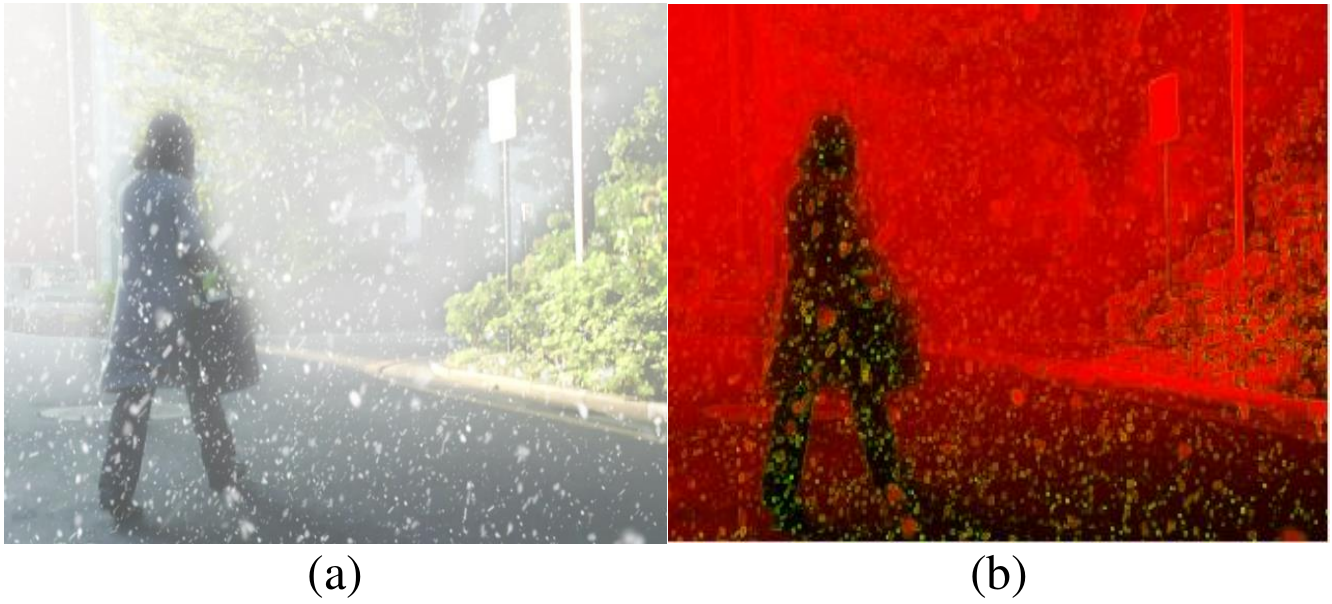}
   \caption{The distribution of degradation interference in the wavelet domain. The left image is a visible light image with degradation interference, while the right image shows the distribution in the wavelet domain.}
   \label{wwb}
\end{figure}

As shown in Figure~\ref{wwb} (b), the red part represents the low-frequency components, and the yellow part represents the high-frequency components. It can be observed that raindrops and snowflakes are mainly distributed in the high-frequency components, while haze is concentrated in the low-frequency components. Low-frequency components can capture global structures and brightness information, thus being suitable for handling large-scale weather interferences such as haze. In contrast, high-frequency components process directional details separately, making them applicable to addressing local interference problems such as rain and snow.
To independently process different degradation types and avoid mutual interference, we adopt different scanning methods for low-frequency and high-frequency components. We use a regular 2D-SSM to scan the low-frequency component, thereby capturing global structural and brightness information, as well as the overall structural and contour information of the image. First, we process the input low-frequency component features $I_{\text{LL}}$  with Layernorm to obtain $F_{\text{LL}}^{FN}$. In one branch, we use Mamba-scan to capture long-range dependencies in the image space. In the other branch, a learnable scaling factor is used to control information from skip connections, preserving original low-frequency information and ensuring the integrity of structural information.
\begin{equation}
I_{\text{LL}}^{out} = F_{\text{LL}}^{FN} * S' + \text{Mamba\_scan}(F_{\text{LL}}^{FN})
\end{equation}
where, $S^{'}$ represents the skip connection, and $\text{Mamba\_scan}$ denotes the regular 2D-SSM scan.

The high-frequency components $I_{\text{HL}}$ , $I_{\text{LH}}$ , $I_{\text{HH}}$  correspond to high-frequency details in different directions. Therefore, we propose Freq-SSM to scan high-frequency components based on this characteristic. Different from the bidirectional scanning of regular 2D-SSM, our Freq-SSM performs bidirectional scanning in three directions, enabling precise directional decomposition. Each direction undergoes independent state-space modeling, enhancing the representation of anisotropic features. Similar to the processing of low-frequency components, we first use Layernorm to process the input high-frequency component features to obtain $F_{\text{High}}$.
\begin{equation}
I_{\text{attn}} = F_{\text{High}} * S' + \text{Freq\_scan}(F_{\text{High}})
\end{equation}
where $I_{\text{attn}}$ and $F_{\text{High}}$ represent ($I_{{\text{LH}}}^{attn}$,  $I_{{\text{HL}}}^{attn}$, $I_{{\text{HH}}}^{attn}$) and ($F_{{\text{LH}}}^{LN}$, $F_{{\text{HL}}}^{LN}$, $F_{{\text{HH}}}^{LN}$), respectively. To enhance the representation capability of different channels, we introduce Channel Attention (CA) to select key channels and avoid channel redundancy. Finally, another tunable scaling factor $S^{'}\in R $ is used in the residual connection to obtain the final output, which can be expressed as:
\begin{equation}
I_{\text{High}}^{out} = \text{CA}(\text{Conv}(LN(I_{\text{attn}}) + S' * I_{\text{attn}})
\end{equation}
The high-frequency and low-frequency components undergo inverse wavelet transform to obtain the complete output $I_{\text{out}}$:
\begin{equation}
I_{\text{out}} = \text{ISWT}(\{I_{{\text{LL}}}^{out},I_{\text{High}}^{out}\})
\end{equation}
\begin{figure}[h]
  \centering
   \includegraphics[width=1\linewidth]{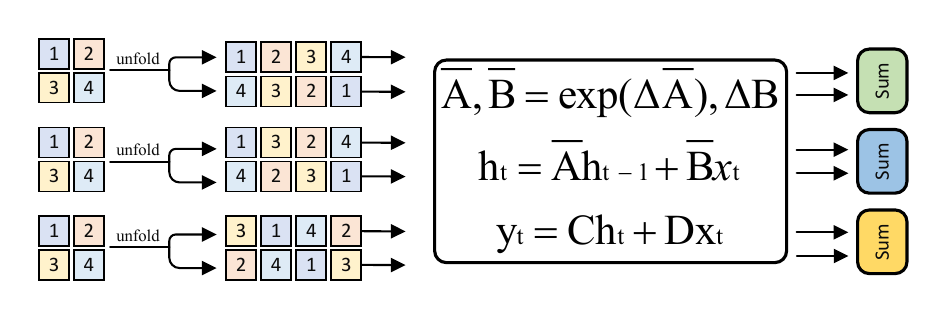}
   \caption{Freq-SSM scanning mechanism}
   \label{fig3}
\end{figure}
\subsubsection{Freq-SSM}
\textbf{Theoretical Motivation.} The design of Freq-SSM is grounded in the intrinsic signal characteristics of compound weather degradations. Unlike general image noise, weather artifacts (e.g., rain streaks, snow) exhibit both frequency specificity and directional anisotropy. While haze primarily manifests as low-frequency contrast degradation, rain and snow are inherently high-frequency signals superimposed with specific directional trends. Standard spatial-domain operations (CNNs) or isotropic scanning mechanisms often fail to disentangle these mixed signals. Compared to other multi-scale representations like Fourier transforms (which lack local spatial localization) or Laplacian pyramids (which lack directional selectivity), the Wavelet Transform is particularly suited for this task due to its Multi-Resolution Analysis (MRA) property. It naturally decouples the signal into distinct subbands with isolated directionality (e.g., LH for vertical, HL for horizontal), providing a sparse and structured basis for subsequent modeling.

\textbf{Preliminaries on Discretized SSM.} To efficiently model these decoupled high-frequency features with global context, we employ the Selective State Space Model (SSM). Unlike Transformers which suffer from quadratic complexity, SSMs offer a global effective receptive field with linear computational complexity. The core involves discretizing the continuous system parameters $(\boldsymbol{\Delta}, \boldsymbol{A}, \boldsymbol{B})$ to discrete parameters $(\overline{\boldsymbol{A}}, \overline{\boldsymbol{B}})$ using the Zero-Order Hold (ZOH) rule:
\begin{equation}
\overline{\boldsymbol{A}} = \exp(\boldsymbol{\Delta}\boldsymbol{A}), \quad \overline{\boldsymbol{B}} = (\boldsymbol{\Delta}\boldsymbol{A})^{-1}(\exp(\boldsymbol{\Delta}\boldsymbol{A}) - \boldsymbol{I}) \cdot \boldsymbol{\Delta}\boldsymbol{B}
\end{equation}
The discretized state equation is then computed recurrently:
\begin{equation}
\boldsymbol{h}_t = \overline{\boldsymbol{A}}\boldsymbol{h}_{t-1} + \overline{\boldsymbol{B}}\boldsymbol{x}_t, \quad \boldsymbol{y}_t = \boldsymbol{C}\boldsymbol{h}_t + \boldsymbol{D}\boldsymbol{x}_t
\end{equation}
Crucially, the parameters $\boldsymbol{\Delta}, \boldsymbol{B}, \boldsymbol{C}$ are learned as input-dependent projections (Selective Scan), allowing the model to dynamically adjust the information flow based on the local intensity of weather artifacts.

\textbf{Directional Alignment Mechanism.} Specifically, Regular 2D-SSM processes inputs by indiscriminately converting them into horizontal and vertical feature representations for selective scanning. This isotropic mechanism creates a fundamental mismatch when applied to high-frequency weather artifacts. For instance, applying a standard multi-directional scan to uni-directional components results in computational redundancy and local feature loss. As shown in Figure~\ref{fig3}, to address this, Freq-SSM aligns the SSM’s scanning trajectory with the dominant direction of each subband (e.g., exclusively vertical scanning for the LH subband). This alignment ensures that the model captures long-range dependencies along the principal axis of the degradation while ignoring irrelevant orthogonal dependencies, thereby maximizing restoration efficiency.

To provide a clear implementation perspective on how the Directional Alignment Mechanism integrates with the Wavelet Transform and SSM operations, we summarize the complete forward process of the proposed module in Algorithm 1.

\begin{algorithm}[t]
\caption{Forward Process of Freq-SSM (WSSB)}
\label{alg:freq_ssm}
\begin{algorithmic}[1]
\Require Feature map $X_{in}$
\Ensure Restored Feature $X_{out}$
\Function{WSSB}{$X_{in}$}
    \State $\{X_{LL}, X_{LH}, X_{HL}, X_{HH}\} \leftarrow \text{DWT}(X_{in})$ \Comment{Wavelet Decomposition}
    \For{each high-freq component $X_{freq} \in \{X_{LH}, X_{HL}, X_{HH}\}$}
        \State $X_{norm} \leftarrow \text{LayerNorm}(X_{freq})$
        \State $X_{scan} \leftarrow \text{DirectionalScan}(X_{norm})$ \Comment{Align with subband direction}
        \State $X_{ssm} \leftarrow \text{SSM}(X_{scan})$ \Comment{Discrete Eq. (2)}
        \State $X_{freq}^{out} \leftarrow X_{freq} + \text{Linear}(X_{ssm})$ \Comment{Residual connection}
    \EndFor
    \State $X_{LL}^{out} \leftarrow \text{RegularSSM}(X_{LL})$ \Comment{Process low-freq base}
    \State $X_{out} \leftarrow \text{IDWT}(\{X_{LL}^{out}, X_{LH}^{out}, X_{HL}^{out}, X_{HH}^{out}\})$
    \State \Return $X_{out}$
\EndFunction
\end{algorithmic}
\end{algorithm}
\begin{figure*}[h]
  \centering
   \includegraphics[width=1.0\linewidth]{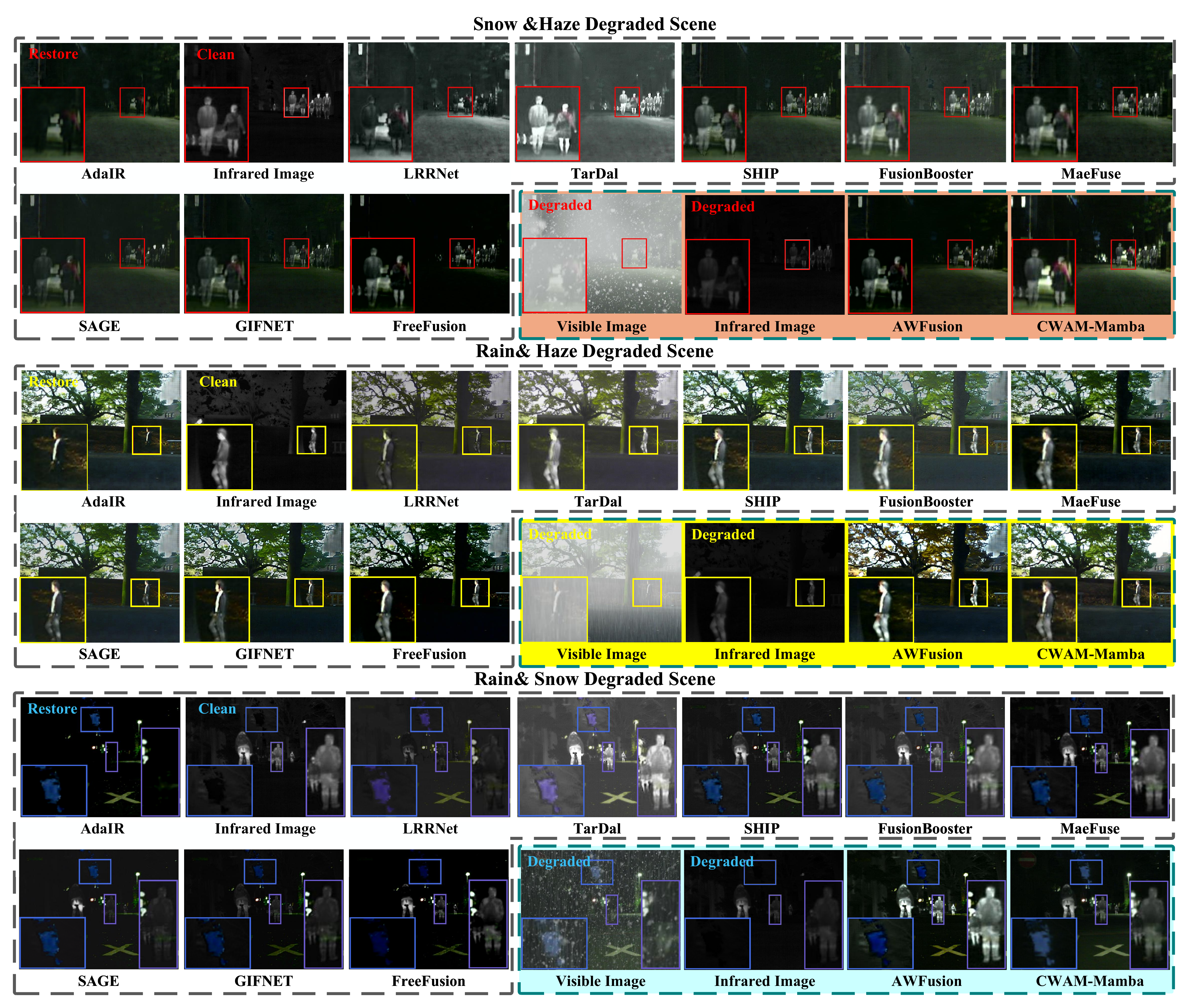}
   \caption{Comparison of fusion results of different methods in three scenarios with composite degradation interference.}
   \label{fig4}
\end{figure*}

\subsubsection{Common Degradation Space Mechanism}
Within WSSB, we further introduce a lightweight Common Degradation Space Mechanism (CDSM) to improve robustness against varying composite degradations.
Instead of treating different degradations independently, CDSM maps degradation-related features into a unified latent space, enabling the network to capture shared patterns across multiple weather types. This mechanism is implemented as a simple residual block with convolution and channel attention, which selectively enhances informative channels while suppressing redundant ones. By integrating this unified degradation representation with the reconstructed wavelet features, WSSB reduces its dependence on specific degradation types and achieves better generalization in compound scenarios.

Given the input feature map $A\in R^{B \times C\times H\times W}$, we first extract basic features through a 3×3 convolution to obtain $A_{in}$, followed by a deep convolution block:
\begin{equation}
A_i = \text{Conv} \left( \sigma \left( \text{Conv}(A_{i-1}) \right) \right) + A_{i-1}
\end{equation}
where, $A_{\text{i}}$ represents the $i^{th}$ convolutional block, . A 3×3 convolution is used to further integrate features processed by multiple convolutional blocks to obtain $A_{\text{mid}}$.
\begin{table*}[h]
\tiny
\label{tablexxx}
\centering
\caption{Comparison of quantitative results of different methods in compound adverse weather scenes. The best scores are in bold, while the second-best scores are in blue.}
\resizebox{\textwidth}{!}{%
\begin{tabular}{c|c|c|c|ccccccccc|c}
\hline
Source                                            & Methods                            & publication                   & Baseline                                    & $Q_{MI}$                  & $Q_{NICE}$                & $Q_{G}$                   & $Q_{M}$                   & $Q_{P}$                   & $Q_{S}$                   & $Q_{CV}$                  & \textit{SSIM}                 & $Q_{abf}$        &Avg.Rank         \\ \hline
\multicolumn{1}{c|}{\multirow{11}{*}{Haze\&Rain}} & \multicolumn{1}{c|}{FusionBooster\citep{FusionBooster}} & \multicolumn{1}{c|}{IJCV'25}  & \multicolumn{1}{c|}{\multirow{9}{*}{AdaIR}} & 0.2606               & 0.8032               & 0.2353               & 0.3804               & 0.1241               & 0.5374               & 1815.6884            & 0.1819               & 0.2811   &8.11            \\
\multicolumn{1}{c|}{}                             & \multicolumn{1}{c|}{GIF-Net\citep{GIFNet}}       & \multicolumn{1}{c|}{CVPR'25}  & \multicolumn{1}{c|}{}                       & 0.2749               & 0.8032               & 0.2943               & 0.3876               & 0.1911               & 0.7261               & 1793.9563            & 0.3100               & 0.3173           &4.78    \\
\multicolumn{1}{c|}{}                             & \multicolumn{1}{c|}{MaeFuse\citep{MaeFuse}}       & \multicolumn{1}{c|}{TIP'25}   & \multicolumn{1}{c|}{}                       & 0.2283               & 0.8031               & 0.2885               & 0.4344               & 0.1464               & 0.7394               & 1200.3360            & 0.2960               & 0.3617         &5.67       \\
\multicolumn{1}{c|}{}                             & \multicolumn{1}{c|}{SAGE\citep{SAGE}}          & \multicolumn{1}{c|}{CVPR'25}  & \multicolumn{1}{c|}{}                       & 0.2318               & 0.8033               & 0.2242               & 0.4389               & 0.1382               & 0.6878               & 1133.0998            & 0.2170               & 0.3143          &6.00     \\
\multicolumn{1}{c|}{}                             & \multicolumn{1}{c|}{SHIP\citep{SHIP}}          & \multicolumn{1}{c|}{CVPR'24}  & \multicolumn{1}{c|}{}                       & 0.2341               & 0.8034               & 0.2793               & \textcolor{blue}{0.4873}               & 0.1519               & 0.6756               & \textcolor{blue}{1058.1555}            & 0.2147               & 0.3843   &  4.56           \\
\multicolumn{1}{c|}{}                             & \multicolumn{1}{c|}{TIM\citep{TIM}}           & \multicolumn{1}{c|}{TPAMI'24} & \multicolumn{1}{c|}{}                       & 0.1569               & 0.8026               & 0.1569               & 0.3264               & 0.0599               & 0.3869               & 1797.5505            & 0.0958               & 0.2368          & 10.78    \\
\multicolumn{1}{c|}{}                             & \multicolumn{1}{c|}{TarDal\citep{TarDal}}        & \multicolumn{1}{c|}{CVPR'22}  & \multicolumn{1}{c|}{}                       & 0.2578               & 0.8034               & 0.2625               & 0.3278               & 0.1606               & 0.4983               & 1166.6283            & 0.1932               & 0.2992        &6.44       \\
\multicolumn{1}{c|}{}                             & \multicolumn{1}{c|}{LRRNet\citep{LRRNet}}        & \multicolumn{1}{c|}{TPAMI'23} & \multicolumn{1}{c|}{}                       & 0.2424               & 0.8034               & 0.2325               & 0.3791               & 0.1324               & 0.6032               & 1264.1777            & 0.0730               & 0.3117          & 7.56    \\
\multicolumn{1}{c|}{}                             & \multicolumn{1}{c|}{FreeFusion\citep{FreeFusion}}    & \multicolumn{1}{c|}{TPAMI'25} & \multicolumn{1}{c|}{}                       & 0.2406               & 0.8030               & 0.1816               & 0.4177               & 0.1565               & 0.4525               & 1144.8330            & 0.1287               & 0.3014        &   7.56     \\ \cline{4-4}
\multicolumn{1}{c|}{}                             & \multicolumn{1}{c|}{AWFusion\citep{AWFusion}}      & \multicolumn{1}{c|}{InfFus'26} & \multicolumn{1}{c|}{-}                      & \textcolor{blue}{0.2749}               & \textcolor{blue}{0.8035}               & \textcolor{blue}{0.3507}               & 0.4092               & \textcolor{blue}{0.2163}               & \textcolor{blue}{0.7753}               & 1257.9992            & \textcolor{blue}{0.3371}               & \textcolor{blue}{0.4096}        & \textcolor{blue}{3.00}       \\
\multicolumn{1}{c|}{}                             & \multicolumn{1}{c|}{CAWM-Mamba}    & \multicolumn{1}{c|}{-}        & \multicolumn{1}{c|}{-}                      & \textbf{0.3604}               & \textbf{0.8054}               & \textbf{0.3761}               & \textbf{0.7426}               & \textbf{ 0.2400}               & \textbf{0.8443}               & \textbf{194.3442}            & \textbf{0.3805}               & \textbf{0.5090}          & \textbf{1.00}    \\ \hline
\multicolumn{1}{l}{}                              & \multicolumn{1}{l}{}               & \multicolumn{1}{l}{}          & \multicolumn{1}{l}{}                        & \multicolumn{1}{l}{} & \multicolumn{1}{l}{} & \multicolumn{1}{l}{} & \multicolumn{1}{l}{} & \multicolumn{1}{l}{} & \multicolumn{1}{l}{} & \multicolumn{1}{l}{} & \multicolumn{1}{l}{} & \multicolumn{1}{l}{} \\ \hline
\multicolumn{1}{c|}{\multirow{11}{*}{Rain\&Snow}} & \multicolumn{1}{c|}{FusionBooster\citep{FusionBooster}} & \multicolumn{1}{c|}{IJCV'25}  & \multicolumn{1}{c|}{\multirow{9}{*}{AdaIR}} & 0.4071               & 0.8050               & 0.3062               & 0.4371               & 0.1817               & 0.6398               & 535.8695             & 0.2469               & 0.3912          & 5.89     \\
\multicolumn{1}{c|}{}                             & \multicolumn{1}{c|}{GIF-Net\citep{GIFNet}}       & \multicolumn{1}{c|}{CVPR'25}  & \multicolumn{1}{c|}{}                       & 0.3907               & 0.8044               & 0.2292               & 0.3561               & 0.1740               & 0.7034               & 497.7260             & 0.2467               & 0.2994            &  7.00  \\
\multicolumn{1}{c|}{}                             & \multicolumn{1}{c|}{MaeFuse\citep{MaeFuse}}       & \multicolumn{1}{c|}{TIP'25}   & \multicolumn{1}{c|}{}                       & 0.3485               & 0.8044               & 0.3091               & 0.4657               & 0.1694               & 0.7376               & 327.2211             & 0.3023               & 0.4129              & 5.44  \\
\multicolumn{1}{c|}{}                             & \multicolumn{1}{c|}{SAGE\citep{SAGE}}          & \multicolumn{1}{c|}{CVPR'25}  & \multicolumn{1}{c|}{}                       & 0.3876               & 0.8051               & 0.2083               & 0.4540               & 0.1770               & \textcolor{blue}{0.7663}               & \textcolor{blue}{270.8444}             & 0.2345               & 0.3361        &  5.56        \\
\multicolumn{1}{c|}{}                             & \multicolumn{1}{c|}{SHIP\citep{SHIP}}          & \multicolumn{1}{c|}{CVPR'24}  & \multicolumn{1}{c|}{}                       & 0.3734               & 0.8052               & 0.3631               & \textcolor{blue}{0.4809}               & 0.2120               & 0.7556               & 303.8411             & 0.3013               & \textcolor{blue}{0.4672}        &  \textcolor{blue}{3.56}       \\
\multicolumn{1}{c|}{}                             & \multicolumn{1}{c|}{TIM\citep{TIM}}           & \multicolumn{1}{c|}{TPAMI'24} & \multicolumn{1}{c|}{}                       & 0.3249               & 0.8039               & 0.1761               & 0.3390               & 0.0935               & 0.4456               & 1438.8557            & 0.1155               & 0.2569          &     10.44  \\
\multicolumn{1}{c|}{}                             & \multicolumn{1}{c|}{TarDal\citep{TarDal}}        & \multicolumn{1}{c|}{CVPR'22}  & \multicolumn{1}{c|}{}                       & 0.3852               & 0.8052               & 0.3182               & 0.3460               & 0.1890               & 0.5123               & 611.4078             & 0.2128               & 0.3510        &  6.89       \\
\multicolumn{1}{c|}{}                             & \multicolumn{1}{c|}{LRRNet\citep{LRRNet}}        & \multicolumn{1}{c|}{TPAMI'23} & \multicolumn{1}{c|}{}                       & \textcolor{blue}{0.4408}               & \textcolor{blue}{0.8056}               & 0.2213               & 0.4116               & 0.1564               & 0.6513               & 750.1812             & 0.0407               & 0.3229        & 7.11         \\
\multicolumn{1}{c|}{}                             & \multicolumn{1}{c|}{FreeFusion\citep{FreeFusion}}    & \multicolumn{1}{c|}{TPAMI'25} & \multicolumn{1}{c|}{}                       & 0.3315               & 0.8040               & 0.1190               & 0.3616               & 0.2111               & 0.2262               & 454.7970             & 0.0781               & 0.2551          & 8.78       \\ \cline{4-4}
\multicolumn{1}{c|}{}                             & \multicolumn{1}{c|}{AWFusion\citep{AWFusion}}      & \multicolumn{1}{c|}{InfFus'26} & \multicolumn{1}{c|}{-}                      & 0.4132               & 0.8053               & \textbf{0.3869}               & 0.3548               & \textbf{0.2534}               & 0.7343               & 571.1523             & \textcolor{blue}{0.3483}               & 0.4309        & 3.89         \\
\multicolumn{1}{c|}{}                             & \multicolumn{1}{c|}{CAWM-Mamba}    & \multicolumn{1}{c|}{-}        & \multicolumn{1}{c|}{-}                      & \textbf{0.4770 }               & \textbf{ 0.8090}               & \textcolor{blue}{0.3826}               & \textbf{0.7994}               & \textcolor{blue}{0.2494}               & \textbf{0.8459}               & \textbf{183.1268 }            & \textbf{0.3910}               & \textbf{0.5143  }          &  \textbf{1.22 }  \\ \hline
\multicolumn{1}{l}{}                              & \multicolumn{1}{l}{}               & \multicolumn{1}{l}{}          & \multicolumn{1}{l}{}                        & \multicolumn{1}{l}{} & \multicolumn{1}{l}{} & \multicolumn{1}{l}{} & \multicolumn{1}{l}{} & \multicolumn{1}{l}{} & \multicolumn{1}{l}{} & \multicolumn{1}{l}{} & \multicolumn{1}{l}{} & \multicolumn{1}{l}{} \\ \hline
\multicolumn{1}{c|}{\multirow{11}{*}{Haze\&Snow}} & \multicolumn{1}{c|}{FusionBooster\citep{FusionBooster}} & \multicolumn{1}{c|}{IJCV'25}  & \multicolumn{1}{c|}{\multirow{9}{*}{AdaIR}} & 0.2580               & 0.8032               & 0.2436               & 0.3806               & 0.1231               & 0.5478               & 1959.6256            & 0.1967               & 0.2943          & 8.00       \\
\multicolumn{1}{c|}{}                             & \multicolumn{1}{c|}{GIF-Net\citep{GIFNet}}       & \multicolumn{1}{c|}{CVPR'25}  & \multicolumn{1}{c|}{}                       & 0.2674               & 0.8032               & 0.2897               & 0.3759               & 0.1737               & 0.7295               & 1922.3891            & 0.3159               & 0.3164           & 5.56    \\
\multicolumn{1}{c|}{}                             & \multicolumn{1}{c|}{MaeFuse\citep{MaeFuse}}       & \multicolumn{1}{c|}{TIP'25}   & \multicolumn{1}{c|}{}                       & 0.2313               & 0.8032               & 0.2923               & 0.4218               & 0.1387               & 0.7419               & 1397.8154            & 0.3087               & 0.3706        &   5.22       \\
\multicolumn{1}{c|}{}                             & \multicolumn{1}{c|}{SAGE\citep{SAGE}}          & \multicolumn{1}{c|}{CVPR'25}  & \multicolumn{1}{c|}{}                       & 0.2395               & 0.8034               & 0.2259               & 0.4307               & 0.1349               & 0.6957               & 1369.7558            & 0.2315               & 0.3204          &  5.78    \\
\multicolumn{1}{c|}{}                             & \multicolumn{1}{c|}{SHIP\citep{SHIP}}          & \multicolumn{1}{c|}{CVPR'24}  & \multicolumn{1}{c|}{}                       & 0.2410               & 0.8034               & 0.2950               & \textcolor{blue}{0.4872}               & 0.1504               & 0.6922               & \textcolor{blue}{1295.2071}            & 0.2398               & \textcolor{blue}{0.4024}          &    3.89  \\
\multicolumn{1}{c|}{}                             & \multicolumn{1}{c|}{TIM\citep{TIM}}           & \multicolumn{1}{c|}{TPAMI'24} & \multicolumn{1}{c|}{}                       & 0.1671               & 0.8027               & 0.1643               & 0.3216               & 0.0629               & 0.3943               & 1957.5747            & 0.1048               & 0.2441          &  10.78    \\
\multicolumn{1}{c|}{}                             & \multicolumn{1}{c|}{TarDal\citep{TarDal}}        & \multicolumn{1}{c|}{CVPR'22}  & \multicolumn{1}{c|}{}                       & 0.2626               & 0.8034               & 0.2793               & 0.3334               & 0.1536               & 0.5041               & 1363.2947            & 0.2059               & 0.3171           &  5.89    \\
\multicolumn{1}{c|}{}                             & \multicolumn{1}{c|}{LRRNet\citep{LRRNet}}        & \multicolumn{1}{c|}{TPAMI'23} & \multicolumn{1}{c|}{}                       & 0.2475               & 0.8034               & 0.2390               & 0.3710               & 0.1310               & 0.5982               & 1561.8435            & 0.0846               & 0.3242          &   7.33    \\
\multicolumn{1}{c|}{}                             & \multicolumn{1}{c|}{FreeFusion\citep{FreeFusion}}    & \multicolumn{1}{c|}{TPAMI'25} & \multicolumn{1}{c|}{}                       & 0.2348               & 0.8030               & 0.1711               & 0.4035               & 0.1383               & 0.4123               & 1428.9506            & 0.1173               & 0.2930          &  8.44  \\ \cline{4-4}
\multicolumn{1}{c|}{}                             & \multicolumn{1}{c|}{AWFusion\citep{AWFusion}}      & \multicolumn{1}{c|}{InfFus'26} & \multicolumn{1}{c|}{-}                      & \textcolor{blue}{0.2858}               & \textcolor{blue}{0.8036}               & \textcolor{blue}{0.3429 }               & 0.3829               & \textcolor{blue}{0.1950 }               & \textcolor{blue}{0.7593 }               & 1482.5806            & \textcolor{blue}{0.3382 }               & 0.3992           & \textcolor{blue}{3.11}    \\
\multicolumn{1}{c|}{}                             & \multicolumn{1}{c|}{CAWM-Mamba}    & \multicolumn{1}{c|}{-}        & \multicolumn{1}{c|}{-}                      & \textbf{0.3675}               & \textbf{0.8056}               & \textbf{0.3772 }               & \textbf{0.7124 }               & \textbf{0.2363}               & \textbf{0.8402}               & \textbf{206.4099 }            & \textbf{ 0.3915}               & \textbf{ 0.5114 }         &   \textbf{1.00 }    \\ \hline
\end{tabular}%
}
\end{table*}

When facing images with different degradation conditions, the channel attention mechanism helps the model focus on key feature channels related to degradation information, thereby better identifying common degradation features:
\begin{equation}
att = \text{Sigmoid} \left( \text{Conv1} \left( \sigma \left( \text{Conv1} \left( \text{Avgpool}(A_{\text{mid}}) \right) \right) \right) \right)
\end{equation}
Then apply it to the feature map:
\begin{equation}
A_{\text{final}} = att * A_{\text{mid}}
\end{equation}
finally, a global residual connection is performed, adding the processed feature map to the initial input feature. This helps preserve the original information in the input features while integrating the processed feature information, enabling the model to better utilize feature information at different levels for image restoration.

Different weather conditions (such as haze, rain, snow, etc.) have varying impacts on images. By introducing weather embeddings obtained from WAPM, image features are adjusted according to different weather conditions, enhancing the model's perception of compound weather interferences like haze, rain, and snow. This enables the model to optimize feature processing strategies by incorporating weather factors during multi-modal image fusion and restoration, improving adaptability and robustness to different weather conditions and making fusion and restoration results more accurate.

\textbf{Weather-Guided Feature Modulation:} To enable adaptive restoration, we introduce a weather-guided modulation mechanism within the WSSB. Specifically, the global weather embedding $F_{embedding} \in \mathbb{R}^{48}$ extracted by WAPM is injected to modulate the reconstructed features via a three-stage post-processing gate. Unlike standard pre-modulation, we first extract global spatial statistics (via Global Average Pooling) from the features refined by both ISWT and the CDSM module and concatenate them with the projected weather embedding. This strategic combination ensures the gating network contextually analyzes the fully restored content alongside the degradation prior.

Subsequently, this combined descriptor passes through a Multi-Layer Perceptron (Linear $\rightarrow$ ReLU $\rightarrow$ Linear $\rightarrow$ Sigmoid) to generate channel-wise calibration weights $w \in [0, 1]^C$. Finally, these weights are applied back to the same refined features via element-wise multiplication ($X_{out} = X_{refined} \odot w$). This operation acts as a dynamic frequency-selection gate, efficiently filtering out residual degradation noise specific to the implicitly learned weather intensity, thereby enhancing the robustness of the final fused output against complex environmental interferences.
\subsection{Loss Functions}
In addition to the new modules, we also introduce a  wavelet loss function to optimize the network training process for better performance in the frequency domain. A one-layer wavelet decomposition is performed on the fused image, and L1 loss is calculated for the low-frequency subband and high-frequency subbands (three directions: $I_{\text{HL}}$ , $I_{\text{LH}}$ , $I_{\text{HH}}$ ). $L_{\text{wavelet}}$ can be expressed as:
\begin{equation}
L_{\text{wavelet}} = L_{\text{low}} + \frac{L_{LH} + L_{HL} + L_{HH}}{3}
\end{equation}
where:
\begin{equation}
L_{ii} = \lVert I_{ii_{\text{fused}}} - I_{ii_{\text{vis}}} \rVert_1
\end{equation}
$L_{\text{ii}}$ represents the component after wavelet decomposition.

Additionally, we introduce other pixel-level loss functions to ensure the network generates high-fidelity fusion results. These loss functions include brightness loss $L_{\text{int}}$, color consistency loss $L_{\text{color}}$, perceptual loss $L_{\text{per}}$, gradient loss $L_{\text{grad}}$, and structural similarity loss $L_{\text{SSIM}}$. $L_{\text{int}}$ can be expressed as:
\begin{equation}
L_{\text{int}} = \frac{1}{HW} \lVert I_{\text{fused}} - \max(I_{\text{vis}}, I_{\text{ir}}) \rVert_1
\end{equation}
where, $I_{\text{vis}}$, $I_{\text{ir}}$ and $I_{\text{fused}}$ respectively represent the clean visible image, the clean infrared image and the fused image.

$L_{\text{color}}$ can be expressed as:
\begin{equation}
L_{\text{color}} = \frac{1}{HW} \lVert T_{CbCr}(I_{\text{fused}}) - T_{CbCr}(I_{\text{vis}}) \rVert_1
\end{equation}
where $T_{\text{CbCr}}$ denotes the transfer function from RGB to CbCr.
\begin{figure*}[h]
  \centering
   \includegraphics[width=1.0\linewidth]{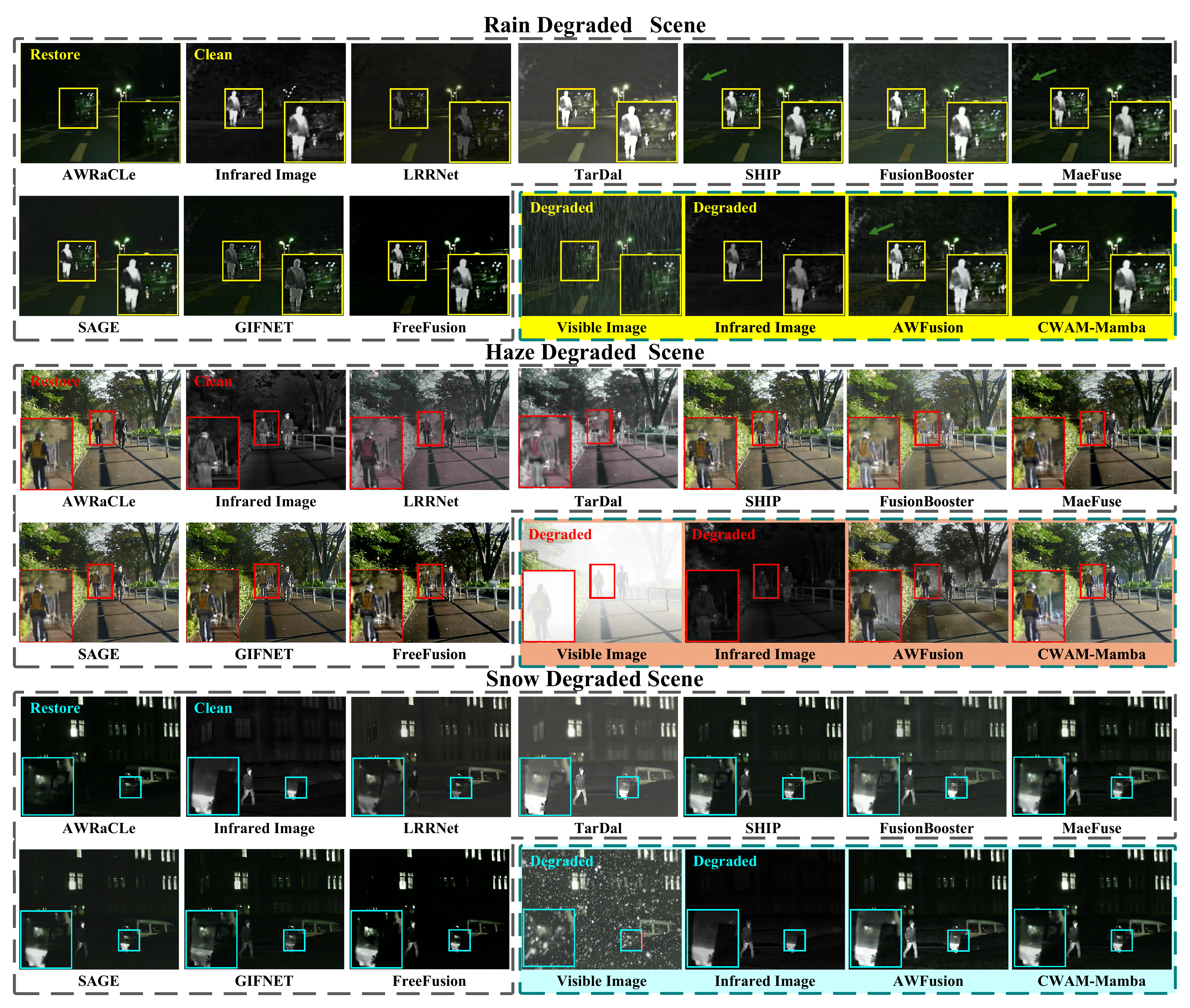}
   \vspace{-6mm}
   \caption{Comparison of fusion results of different methods in three scenarios with single degradation interference.}
   \label{fig5}
\end{figure*}

$L_{\text{per}}$ can be expressed as:
\begin{equation}
L_{\text{per}} = \frac{1}{L} \sum_l \lVert \phi_l(I_{\text{fused}}) - \phi_l(I_{\text{vis}}) \rVert_1
\end{equation}
where $\phi_l$ denotes the feature layer of MobileNetV3.

$L_{\text{grad}}$ can be expressed as:
\begin{equation}
L_{\text{grad}} = \frac{1}{HW} \lVert \nabla I_{\text{fused}} - \max(\nabla I_{\text{vis}}, \nabla I_{\text{ir}}) \rVert_1
\end{equation}

$L_{\text{SSIM}}$ can be expressed as:
\begin{equation}
L_{\text{SSIM}} = 1 - \text{SSIM}(I_{\text{vis}}, I_{\text{ir}}, I_{\text{fused}})
\end{equation}
where $\text{SSIM}(\cdot)$ denotes the structural similarity measurement operator.

The overall loss function is expressed as:
\begin{equation}
L_{\text{Total}} = \alpha L_{\text{wavelet}} + L_{\text{color}} + L_{\text{per}} + L_{\text{grad}} + L_{\text{int}} + L_{\text{SSIM}}
\end{equation}
where $\alpha$ is a learnable scalar initialized to 0.5. It serves as a dynamic weighting factor for the total wavelet loss and is optimized jointly with the network parameters. This adaptive design allows the model to automatically balance the importance of frequency-domain consistency relative to spatial-domain constraints during the training process.
\section{Experiment}
In this section, we first introduce the experimental details and relevant configurations. Then, the effectiveness of the proposed method is evaluated through qualitative and quantitative comparisons. Finally, ablation experiments are conducted.
\subsection{Experimental Setup.}
\subsubsection{Training Details.}
Our proposed CAWM-Mamba is trained using two Nvidia GeForce RTX 3090 GPUs. During training, the image cropping size gradually expands from 64 to 144. We use the Adam optimizer with default parameters to optimize our algorithm, setting the initial learning rate to 0.0001. In terms of network configuration, our WSSBs adopt a U-Net architecture, with the block configuration of each layer set to [8,10,10,12,10,10,8] to 
\begin{table*}[h]
\centering
\tiny
\caption{Comparison of quantitative results of different methods in single adverse weather scenes. The best scores are in bold, while the second-best scores are in blue.}
\label{table2}
\resizebox{\textwidth}{!}{%
\begin{tabular}{ccccccccccccc|c}
\hline
\multicolumn{1}{c|}{Source}                 & \multicolumn{1}{c|}{Methods}       & \multicolumn{1}{c|}{Publication} & \multicolumn{1}{c|}{Baseline}                 & $Q_{MI}$                  & $Q_{NICE}$                & $Q_{G}$                   & $Q_{M}$                   & $Q_{P}$                   & $Q_{S}$                   & $Q_{CV}$                  &\textit{SSIM}                  & $Q_{abf}$    &Avg.Rank              \\ \hline
\multicolumn{1}{c|}{\multirow{11}{*}{Rain}} & \multicolumn{1}{c|}{FusionBooster\citep{FusionBooster}} & \multicolumn{1}{c|}{IJCV'25}     & \multicolumn{1}{c|}{\multirow{9}{*}{AWRaCLe}} & 0.3065               & 0.8037               & 0.2830               & 0.4086               & 0.1646               & 0.6247               & 812.3392             & 0.2473               & 0.3517         &  8.33    \\
\multicolumn{1}{c|}{}                       & \multicolumn{1}{c|}{GIF-Net\citep{GIFNet}}       & \multicolumn{1}{c|}{CVPR'25}     & \multicolumn{1}{c|}{}                         & 0.2927               & 0.8036               & 0.3037               & 0.3712               & 0.2071               & 0.8060               & 664.2364             & 0.3499               & 0.3605           &  6.67  \\
\multicolumn{1}{c|}{}                       & \multicolumn{1}{c|}{MaeFuse\citep{MaeFuse}}       & \multicolumn{1}{c|}{TIP'25}      & \multicolumn{1}{c|}{}                         & 0.3126               & 0.8041               & 0.3372               & 0.4446               & 0.1888               & \textcolor{blue}{0.8337}               & 580.1806             &\textcolor{blue}{ 0.3680 }             & 0.4280         &    4.67   \\
\multicolumn{1}{c|}{}                       & \multicolumn{1}{c|}{SAGE\citep{SAGE}}          & \multicolumn{1}{c|}{CVPR'25}     & \multicolumn{1}{c|}{}                         & \textcolor{blue}{0.4134}               & \textcolor{blue}{0.8062}               & 0.2304               & 0.5201               & 0.1872               & 0.7899               & 462.2776             & 0.2809               & 0.3729         &    4.89   \\
\multicolumn{1}{c|}{}                       & \multicolumn{1}{c|}{SHIP\citep{SHIP}}          & \multicolumn{1}{c|}{CVPR'24}     & \multicolumn{1}{c|}{}                         & 0.3966               & 0.8056               & \textcolor{blue}{0.3777}               & \textcolor{blue}{0.6662}               & \textcolor{blue}{0.2299}               & 0.8049               & \textcolor{blue}{452.2582}             & 0.3347               & \textcolor{blue}{0.4923}       &      \textcolor{blue}{3.11}    \\
\multicolumn{1}{c|}{}                       & \multicolumn{1}{c|}{TIM\citep{TIM}}           & \multicolumn{1}{c|}{TPAMI'24}    & \multicolumn{1}{c|}{}                         & 0.2659               & 0.8039               & 0.1877               & 0.3498               & 0.0780               & 0.4702               & 922.9942             & 0.1190               & 0.2774          & 10.33    \\
\multicolumn{1}{c|}{}                       & \multicolumn{1}{c|}{TarDal\citep{TarDal}}        & \multicolumn{1}{c|}{CVPR'22}     & \multicolumn{1}{c|}{}                         & 0.3327               & 0.8045               & 0.3173               & 0.3532               & 0.1914               & 0.5589               & 797.0368             & 0.2462               & 0.3428         &  7.22     \\
\multicolumn{1}{c|}{}                       & \multicolumn{1}{c|}{LRRNet\citep{LRRNet}}        & \multicolumn{1}{c|}{TPAMI'23}    & \multicolumn{1}{c|}{}                         & 0.4035               & 0.8057               & 0.2336               & 0.4143               & 0.1721               & 0.6849               & 805.5428             & 0.0486               & 0.3400       &    7.11     \\
\multicolumn{1}{c|}{}                       & \multicolumn{1}{c|}{FreeFusion\citep{FreeFusion}}    & \multicolumn{1}{c|}{TPAMI'25}    & \multicolumn{1}{c|}{}                         & 0.3183               & 0.8038               & 0.1817               & 0.3881               & 0.1695               & 0.3866               & 642.0901             & 0.1384               & 0.3133        & 8.89        \\ \cline{4-4}
\multicolumn{1}{c|}{}                       & \multicolumn{1}{c|}{AWFusion\citep{AWFusion}}      & \multicolumn{1}{c|}{InfFus'26}    & \multicolumn{1}{c|}{-}                         & 0.3509               & 0.8054               & 0.3334               & 0.6428               & 0.1962               & 0.8335               & 476.3436             & 0.3528               & 0.4618       &  3.78        \\
\multicolumn{1}{c|}{}                       & \multicolumn{1}{c|}{CAWM-Mamba}    & \multicolumn{1}{c|}{-}           & \multicolumn{1}{c|}{-}                         & \textbf{0.4310 }               & \textbf{ 0.8074}               & \textbf{0.3898}               & \textbf{0.7575}               & \textbf{0.2363}               & \textbf{0.8448}               & \textbf{451.1764}             & \textbf{0.3998}               & \textbf{0.5024 }      &  \textbf{1.00 }      \\ \hline
\multicolumn{1}{l}{}                        & \multicolumn{1}{l}{}               & \multicolumn{1}{l}{}             & \multicolumn{1}{l}{}                           & \multicolumn{1}{l}{} & \multicolumn{1}{l}{} & \multicolumn{1}{l}{} & \multicolumn{1}{l}{} & \multicolumn{1}{l}{} & \multicolumn{1}{l}{} & \multicolumn{1}{l}{} & \multicolumn{1}{l}{} & \multicolumn{1}{l}{} \\ \hline
\multicolumn{1}{c|}{\multirow{11}{*}{Haze}} & \multicolumn{1}{c|}{FusionBooster\citep{FusionBooster}} & \multicolumn{1}{c|}{IJCV'25}     & \multicolumn{1}{c|}{\multirow{9}{*}{AWRaCLe}} & 0.2589               & 0.8036               & 0.3160               & 0.3760               & 0.2237               & 0.6707               & 829.1121             & 0.3151               & 0.3794         &9.44     \\
\multicolumn{1}{c|}{}                       & \multicolumn{1}{c|}{GIF-Net\citep{GIFNet}}       & \multicolumn{1}{c|}{CVPR'25}     & \multicolumn{1}{c|}{}                         & 0.2621               & 0.8036               & 0.3477               & 0.3375               & 0.2681               & 0.7223               & 657.4543             & 0.3715               & 0.4015         &   8.44   \\
\multicolumn{1}{c|}{}                       & \multicolumn{1}{c|}{MaeFuse\citep{MaeFuse}}       & \multicolumn{1}{c|}{TIP'25}      & \multicolumn{1}{c|}{}                         & 0.2476               & 0.8038               & 0.3712               & 0.3943               & 0.2636               & 0.7486               & 586.8140             & 0.3768               & 0.4490        &  7.00      \\
\multicolumn{1}{c|}{}                       & \multicolumn{1}{c|}{SAGE\citep{SAGE}}          & \multicolumn{1}{c|}{CVPR'25}     & \multicolumn{1}{c|}{}                         & 0.2920               & 0.8046               & 0.3948               & 0.5019               & 0.3396               &  \textcolor{blue}{0.7771}               & 398.6333             & 0.3781               & 0.5072        &  3.44     \\
\multicolumn{1}{c|}{}                       & \multicolumn{1}{c|}{SHIP\citep{SHIP}}          & \multicolumn{1}{c|}{CVPR'24}     & \multicolumn{1}{c|}{}                         & 0.2914               & 0.8047               & \textcolor{blue}{0.4479}               &  \textcolor{blue}{0.5850}               & 0.3557               & 0.7702               &  \textcolor{blue}{377.2266 }             & 0.3730               & \textcolor{blue}{ 0.5589 }       &   \textcolor{blue}{3.00}     \\
\multicolumn{1}{c|}{}                       & \multicolumn{1}{c|}{TIM\citep{TIM}}           & \multicolumn{1}{c|}{TPAMI'24}    & \multicolumn{1}{c|}{}                         & 0.2340               & 0.8037               & 0.2557               & 0.3646               & 0.1479               & 0.6102               & 581.2153             & 0.2362               & 0.3683        & 10.11        \\
\multicolumn{1}{c|}{}                       & \multicolumn{1}{c|}{TarDal\citep{TarDal}}        & \multicolumn{1}{c|}{CVPR'22}     & \multicolumn{1}{c|}{}                         & 0.2833               & 0.8042               & 0.3545               & 0.3467               & 0.2724               & 0.6979               & 488.7347             & 0.3536               & 0.3829        &  6.89     \\
\multicolumn{1}{c|}{}                       & \multicolumn{1}{c|}{LRRNet\citep{LRRNet}}        & \multicolumn{1}{c|}{TPAMI'23}    & \multicolumn{1}{c|}{}                         & \textcolor{blue}{0.3038}               & \textcolor{blue}{0.8048}               & 0.3500               & 0.4114               & 0.3026               & 0.7242               & 559.1792             & 0.2401               & 0.4411           &   5.78 \\
\multicolumn{1}{c|}{}                       & \multicolumn{1}{c|}{FreeFusion\citep{FreeFusion}}    & \multicolumn{1}{c|}{TPAMI'25}    & \multicolumn{1}{c|}{}                         & 0.2662               & 0.8039               & 0.3370               & 0.3997               & 0.3074               & 0.5318               & 522.8681             & 0.2600               & 0.4537        &    7.11    \\ \cline{4-4}
\multicolumn{1}{c|}{}                       & \multicolumn{1}{c|}{AWFusion\citep{AWFusion}}      & \multicolumn{1}{c|}{InfFus'26}    & \multicolumn{1}{c|}{-}                         & 0.2639               & 0.8042               & 0.4283               & 0.5471               &  \textcolor{blue}{0.3587}               & 0.7738               & 462.8585             &  \textcolor{blue}{0.3901 }             & 0.5322        &  3.56       \\
\multicolumn{1}{c|}{}                       & \multicolumn{1}{c|}{CAWM-Mamba}    & \multicolumn{1}{c|}{-}           & \multicolumn{1}{c|}{-}                         & \textbf{0.5008}               & \textbf{0.8112}               & \textbf{0.5394}               & \textbf{0.8821}               & \textbf{0.4857 }               & \textbf{0.8370}               & \textbf{159.6150}             & \textbf{ 0.4705 }             & \textbf{0.6344}       & \textbf{1.00}        \\ \hline
\multicolumn{1}{l}{}                        & \multicolumn{1}{l}{}               & \multicolumn{1}{l}{}             & \multicolumn{1}{l}{}                           & \multicolumn{1}{l}{} & \multicolumn{1}{l}{} & \multicolumn{1}{l}{} & \multicolumn{1}{l}{} & \multicolumn{1}{l}{} & \multicolumn{1}{l}{} & \multicolumn{1}{l}{} & \multicolumn{1}{l}{} & \multicolumn{1}{l}{} \\ \hline
\multicolumn{1}{c|}{\multirow{11}{*}{Snow}} & \multicolumn{1}{c|}{FusionBooster\citep{FusionBooster}} & \multicolumn{1}{c|}{IJCV'25}     & \multicolumn{1}{c|}{\multirow{9}{*}{AWRaCLe}} & 0.3489               & 0.8046               & 0.3149               & 0.3720               & 0.2151               & 0.6415               & 831.4537             & 0.2893               & 0.3933       &8.56        \\
\multicolumn{1}{c|}{}                       & \multicolumn{1}{c|}{GIF-Net\citep{GIFNet}}       & \multicolumn{1}{c|}{CVPR'25}     & \multicolumn{1}{c|}{}                         & 0.3385               & 0.8045               & 0.3242               & 0.3289               & 0.2476               & 0.7690               & 485.5636             & 0.3704               & 0.3854        &   7.89     \\
\multicolumn{1}{c|}{}                       & \multicolumn{1}{c|}{MaeFuse\citep{MaeFuse}}       & \multicolumn{1}{c|}{TIP'25}      & \multicolumn{1}{c|}{}                         & 0.3481               & 0.8050               & 0.3617               & 0.4017               & 0.2451               & 0.7994               & 439.4400             &  \textcolor{blue}{0.3910}               & 0.4597           &5.33    \\
\multicolumn{1}{c|}{}                       & \multicolumn{1}{c|}{SAGE\citep{SAGE}}          & \multicolumn{1}{c|}{CVPR'25}     & \multicolumn{1}{c|}{}                         & 0.4657               & 0.8082               & 0.3107               & 0.4720               & 0.2815               & 0.7908               & 265.9750             & 0.3445               & 0.4556           &  4.56  \\
\multicolumn{1}{c|}{}                       & \multicolumn{1}{c|}{SHIP\citep{SHIP}}          & \multicolumn{1}{c|}{CVPR'24}     & \multicolumn{1}{c|}{}                         & 0.4628               & 0.8086               & \textcolor{blue}{ 0.4185  }             & \textcolor{blue}{ 0.6150 }             & \textcolor{blue}{ 0.3159}               & 0.7940               &  \textcolor{blue}{236.4233}             & 0.3692               &  \textcolor{blue}{0.5429}        &   \textcolor{blue}{2.89}     \\
\multicolumn{1}{c|}{}                       & \multicolumn{1}{c|}{TIM\citep{TIM}}           & \multicolumn{1}{c|}{TPAMI'24}    & \multicolumn{1}{c|}{}                         & 0.2848               & 0.8044               & 0.2233               & 0.3366               & 0.1346               & 0.5196               & 735.1831             & 0.1819               & 0.3246        &10.44        \\
\multicolumn{1}{c|}{}                       & \multicolumn{1}{c|}{TarDal\citep{TarDal}}        & \multicolumn{1}{c|}{CVPR'22}     & \multicolumn{1}{c|}{}                         & 0.4005               & 0.8062               & 0.3405               & 0.3308               & 0.2587               & 0.6080               & 560.5218             & 0.2974               & 0.3784        &   7.33    \\
\multicolumn{1}{c|}{}                       & \multicolumn{1}{c|}{LRRNet\citep{LRRNet}}        & \multicolumn{1}{c|}{TPAMI'23}    & \multicolumn{1}{c|}{}                         & 0.4384               & 0.8071               & 0.2841               & 0.3735               & 0.2547               & 0.7009               & 657.3085             & 0.1551               & 0.3967        &  7.44     \\
\multicolumn{1}{c|}{}                       & \multicolumn{1}{c|}{FreeFusion\citep{FreeFusion}}    & \multicolumn{1}{c|}{TPAMI'25}    & \multicolumn{1}{c|}{}                         & 0.3579               & 0.8049               & 0.2491               & 0.3811               & 0.2576               & 0.4557               & 372.0036             & 0.2004               & 0.3977        &  7.56     \\ \cline{4-4}
\multicolumn{1}{c|}{}                       & \multicolumn{1}{c|}{AWFusion\citep{AWFusion}}     & \multicolumn{1}{c|}{InfFus'26}    & \multicolumn{1}{c|}{-}                         &\textcolor{blue}{0.4668}               &  \textcolor{blue}{0.8094}               & 0.3552               & 0.5535               & 0.2724               & \textcolor{blue}{ 0.8119 }             & 270.1595             & 0.3734               & 0.4937        &   3.00   \\
\multicolumn{1}{c|}{}                       & \multicolumn{1}{c|}{CAWM-Mamba}    & \multicolumn{1}{c|}{-}           & \multicolumn{1}{c|}{-}                         & \textbf{0.5322}               & \textbf{0.8117}               & \textbf{0.4331}               & \textbf{0.7455}               & \textbf{ 0.3490  }             & \textbf{0.8365}               & \textbf{211.1089}             & \textbf{0.4323}               & \textbf{0.5660 }        & \textbf{1.00}     \\ \hline
\end{tabular}%
} 
\end{table*}
balance performance and parameter efficiency. The number of feature channels in WSSBs gradually increases from 48 to 384 and then decreases to 48.
\begin{figure*}[h]
  \centering
   \includegraphics[width=1.0\linewidth]{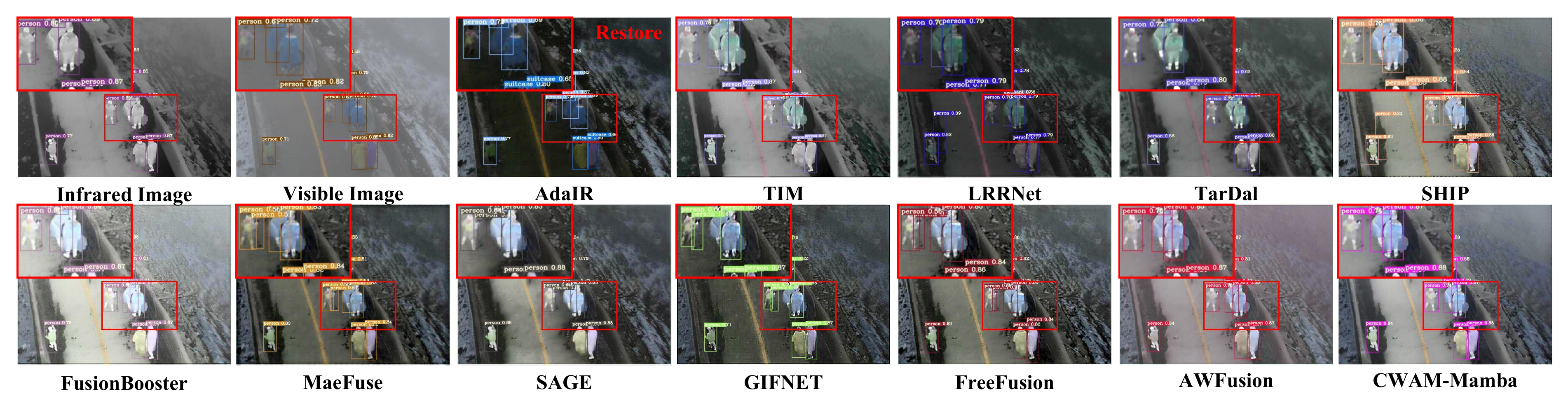}
   \caption{Comparison of fusion results of different methods in real-world scenarios }
   \label{fig7}
\end{figure*}
\begin{figure*}[h]
  \centering
   \includegraphics[width=1.0\linewidth]{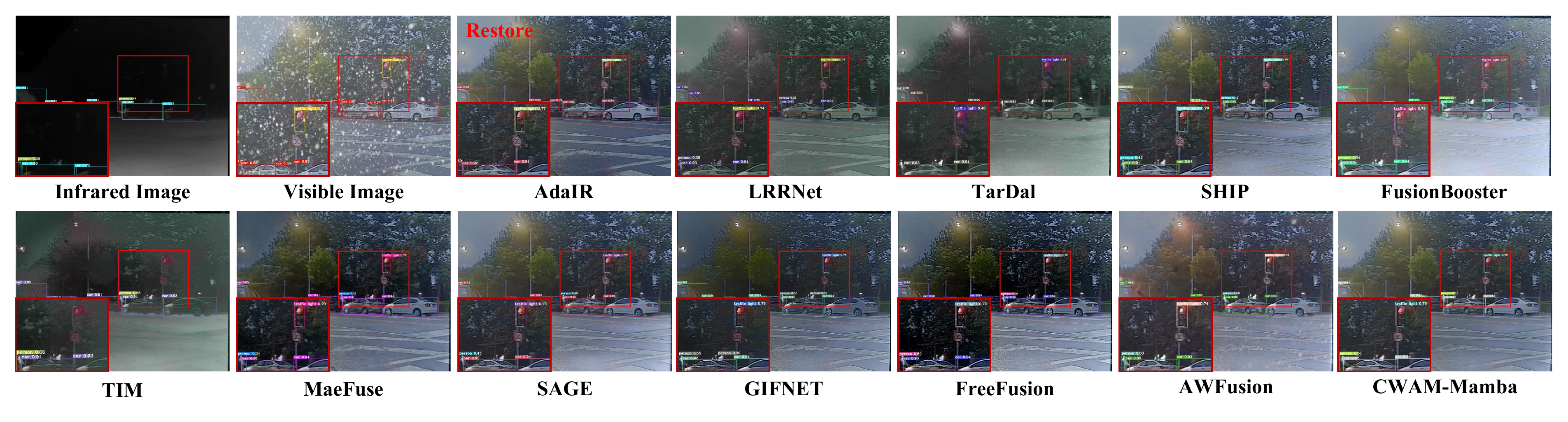}
   \caption{Detection comparisons of different methods. }
   \label{fig9}
\end{figure*}
\begin{table*}[h]
\footnotesize
\caption{Comparing quantitative results of different methods in object detection task.  The best scores are in bold, while the second-best scores are in blue.}
\centering
\label{table4}
\begin{tabular}{c|c|c|cccccccc}
\hline
Methods       & Publication & Baseline                 & People & Car   & Bus   & Lamp  & Motorcycle & Truck & AP@0.5 & AP@{[}0.5:0.95{]} \\ \hline
VI &- &- &0.26 &0.767 &0.715 &0.554 &0.538 &0.592 &0.571 &0.339 \\
IR &- &- &0.605 &0.619 &0.319 &0.013 &0.299 &0.227 &0.347 &0.203 
\\ \cline{3-3}
FusionBooster\citep{FusionBooster} & IJCV'25     & \multirow{9}{*}{AdaIR} & 0.799  & 0.887 & 0.846 & 0.681 & 0.666      & 0.775 & 0.776  & 0.484                \\
GIF-Net\citep{GIFNet}       & CVPR'25     &                          & 0.795  & 0.903 & \textcolor{blue}{0.922} & 0.772 & 0.672      & 0.792 & 0.809  & 0.506                \\
MaeFuse\citep{MaeFuse}       & TIP'25      &                          & 0.799  & 0.885 & 0.904 & 0.720  & 0.669      & 0.798 & 0.796  & 0.503                \\
SAGE\citep{SAGE}          & CVPR'25     &                          & 0.806  & 0.903 & 0.908 & \textcolor{blue}{0.782} & 0.657      & \textbf{0.810}  & 0.811  & 0.516                \\
SHIP\citep{SHIP}          & CVPR'24     &                          & 0.802  & 0.899 & 0.899 & 0.768 & 0.692      & 0.770  & 0.805  & 0.503                \\
TIM\citep{TIM}           & TPAMI'24    &                          & 0.661  & 0.741 & 0.582 & 0.105 & 0.377      & 0.446 & 0.485  & 0.293                \\
TarDal\citep{TarDal}        & CVPR'22     &                          & 0.772  & 0.864 & 0.856 & 0.551 & 0.629      & 0.689 & 0.727  & 0.449                \\
LRRNet\citep{LRRNet}        & TPAMI'23    &                          & 0.759  & 0.894 & 0.906 & 0.731 & 0.700        & 0.776 & 0.794  & 0.496                \\
FreeFusion\citep{FreeFusion}    & TPAMI'25    &                          & \textcolor{blue}{0.809}  & \textcolor{blue}{0.906} & \textbf{0.926} & 0.765 & 0.691      & \textcolor{blue}{0.809} & \textcolor{blue}{0.817}  & \textcolor{blue}{0.520}                 \\ \cline{3-3}
AWFusion\citep{AWFusion}      & InfFus'26    & -                        & 0.800  & 0.888 & 0.869 & 0.668 & \textcolor{blue}{0.717}      & 0.742 & 0.781  & 0.488                \\
CAWM-Mamba    & -           & -                        & \textbf{0.822}  & \textbf{0.911} & 0.909 &\textbf{ 0.800}   &\textbf{ 0.713 }     & 0.802 & \textbf{0.826}  & \textbf{0.528}                \\ \hline
\end{tabular}
\end{table*}
\begin{figure*}[h]
  \centering
   \includegraphics[width=1.0\linewidth]{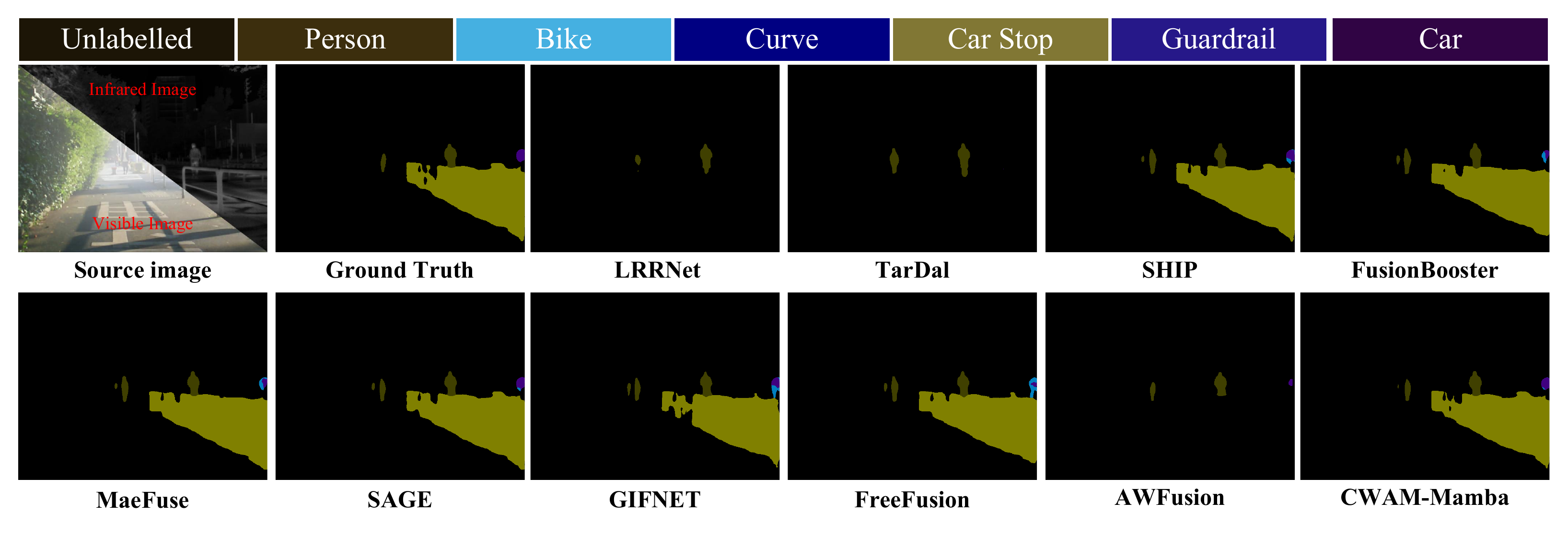}
   \caption{Segmentation comparisons of different methods. }
   \label{fig8}
\end{figure*}
\begin{figure*}[h]
  \centering
   \includegraphics[width=1.0\linewidth]{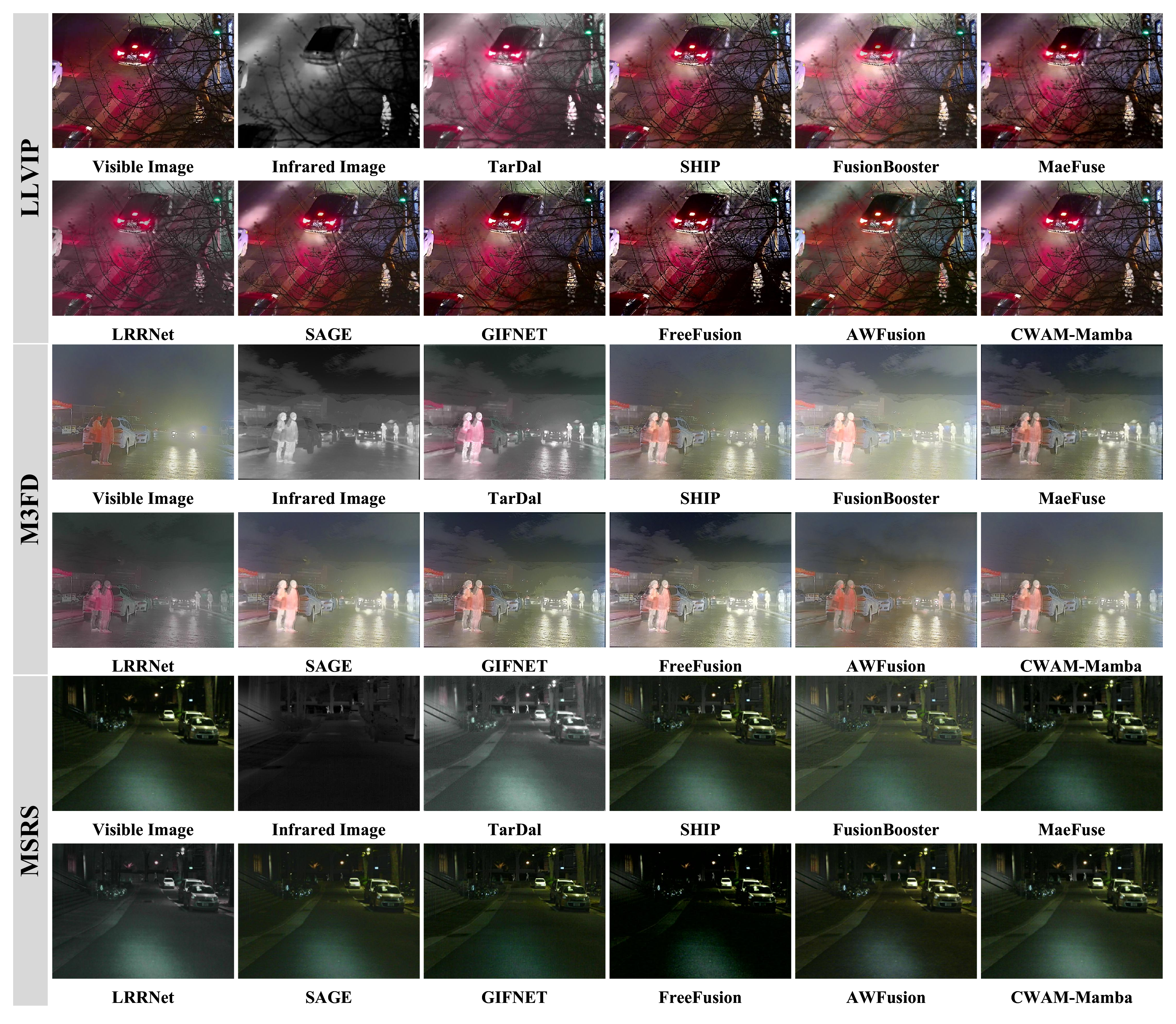}
   \caption{Comparison of fusion results of different methods under three standard scenarios }
   \label{fig6}
\end{figure*}
\begin{table*}[h]
\footnotesize
\caption{Comparing quantitative results of different methods in semantic segmentation task. The best scores are in bold, while the second-best scores are in blue.}
\label{table5}
\centering
\begin{tabular}{c|c|cccccccccc}
\hline
Methods                            & Baseline               & unlabelled & car   & person & bike  & curve & car\_stop & guardrail & color\_cone & bump  & mIoU   \\ \hline

\multicolumn{1}{c|}{VI} &- & 96.71    & 74.57  &25.61  & 49.03  & 38.92  & 60.95  & 70.2  & 52.99 & 68.88 & 59.76     \\

\multicolumn{1}{c|}{IR} & - & 96.12    & 61.17  &69.77  & 23.08  & 34.16  & 23.67  & 0.0  & 20.63 & 27.39 & 39.55    \\ \cline{2-2}

\multicolumn{1}{c|}{FusionBooster\citep{FusionBooster}} & \multirow{9}{*}{AdaIR} & 98.22      & \textcolor{blue}{87.33} & \textbf{72.95}  & 63.76 & 56.12 & 68.88     & \textcolor{blue}{88.35}     & 59.68       & 66.15 & 72.52 \\
\multicolumn{1}{c|}{GIF-Net\citep{GIFNet}}       &                        & 98.17      & 86.65 & 70.63  & 65.23 & 55.32 & 67.77     & 75.67     & 61.38       & 64.67 & 71.72 \\
\multicolumn{1}{c|}{MaeFuse\citep{MaeFuse}}       &                        & 98.30       & \textbf{87.72} & \textcolor{blue}{72.13}  & \textcolor{blue}{66.45} & \textbf{61.91} & \textcolor{blue}{69.2}      & 82.94     & 61.32       & 74.9  & \textcolor{blue}{74.99} \\
\multicolumn{1}{c|}{SAGE\citep{SAGE}}          &                        & 98.11      & 86.21 & 67.56  & 64.68 & 52.15 & 66.84     & \textbf{88.95}     & 60.05       & 73.34 & 72.21 \\
\multicolumn{1}{c|}{SHIP\citep{SHIP}}          &                        & 98.23      & 86.87 & 71.32  & 65.36 & 58.21 & 66.48     & 83.63     & \textbf{68.66}       & \textcolor{blue}{78.78} & 73.5  \\
\multicolumn{1}{c|}{TIM\citep{TIM}}           &                        & 97.29      & 80.63 & 63.69  & 57.84 & 24.12 & 51.36     & 44.78     & 55.14       & 24.47 & 55.39 \\
\multicolumn{1}{c|}{TarDal\citep{TarDal}}        &                        & 97.61      & 84.73 & 67.92  & 58.44 & 48.25 & 43.19     & 71.86     & 23.09       & 30.88 & 58.44 \\
\multicolumn{1}{c|}{LRRNet\citep{LRRNet}}        &                        & 97.52      & 83.77 & 59.03  & 59.68 & 42.34 & 41.23     & 75.32     & 32.25       & 45.51 & 59.63 \\
\multicolumn{1}{c|}{FreeFusion\citep{FreeFusion}}    &                        & \textbf{98.81}      & 85.54 & 71.56  & 62.83 & 45.57 & 63.22     & 88.11     & 54.75       & 64.86 & 69.61 \\ \cline{2-2}
\multicolumn{1}{c|}{AWFusion\citep{AWFusion}}      & -                      & 97.90       & 83.80  & 66.43  & 61.33 & 53.91 & 64.17     & 69.84     & 54.95       & 63.85 & 68.38 \\
\multicolumn{1}{c|}{CAWM-Mamba}    & -                      & \textcolor{blue}{98.30}       & 86.93 & 69.24  & \textbf{68.91} & \textcolor{blue}{59.42} & \textbf{72.30}      & 85.58     & \textcolor{blue}{63.52}       & \textbf{79.63} & \textbf{75.98} \\ \hline
\end{tabular}
\end{table*}
\subsubsection{Datasets.}
We use the AWMM-100K\citep{AWFusion} dataset as the dataset for this experiment. AWMM-100K contains a total of 100,000 real image pairs and synthetic sample image pairs, providing a comprehensive benchmark for evaluating the fusion capabilities of different methods under severe weather conditions. We constructed the datasets with a strict protocol to avoid data leakage. For the single-degradation dataset, we extracted 700 images for each weather type (rain, haze, snow) from AWMM-100K to serve purely as the training set. Additionally, we selected 50 separate images for each degradation type as a distinct test set, ensuring zero scene overlap with the training data. Similarly, for the compound-degradation dataset, we extracted 1,000 images for each type (rain\&haze, rain\&snow and haze\&snow) exclusively for training, and selected 50 separate images for testing. Crucially, the test scenes are completely distinct from the training scenes, and the test sets for compound degradations are also independent from the single-degradation test set. This rigorous separation ensures that the reported results reflect the model's true robustness to unseen adverse weather conditions. To verify that our method also has superior performance in ideal scenarios, we also conducted experiments on three standard datasets (LLVIP\citep{LLVIP}, MSRS\citep{MSRS} and M3FD\citep{M3FD}) using clean multi-modal image pairs.
\subsubsection{Compared Methods and Metrics.}
To verify the superiority of CAWM-Mamba, we compare the proposed method with several state-of-the-art methods in the above datasets through qualitative and quantitative comparisons. The methods compared include LRRNet\citep{LRRNet}, TarDal\citep{TarDal}, SHIP\citep{SHIP}, FusionBooster\citep{FusionBooster}, MaeFuse\citep{MaeFuse}, GIFNet\citep{GIFNet}, SAGE\citep{SAGE}, FreeFusion\citep{FreeFusion}, and AWFusion\citep{AWFusion}. Among them, except for AWFusion, the remaining methods are not designed for complex scenarios. Therefore, these 8 methods use the visible images restored by two baselines (AwaCLe\citep{AWRaCLe}, AdaIR\citep{cui2025adair}) and clean infrared images. AWFusion and our CAWM-Mamba, designed for severe weather scenarios, directly use weather-degraded visible and infrared images for fusion.

The metrics we use include Normalized Mutual Metric $Q_{\text{MI}}$, Nonlinear Correlation Metric $Q_{\text{NICE}}$, Gradient-Based Metric $Q_{\text{G}}$, Multiscale Scheme Metric $Q_{\text{M}}$, Phase Congruency Metric $Q_{\text{P}}$, Piella’s Metric $Q_{\text{S}}$, Chen-Varsheny Metric $Q_{\text{CV}}$, Structural Similarity-Based Metric $\text{SSIM}$, and Quality Assessment-Based Metric $Q_{\text{abf}}$.
For $Q_{\text{CV}}$, a lower value indicates higher fusion image quality; for the remaining metrics, a higher value indicates higher fusion quality.
\subsection{Experiment Result}
\subsubsection{Image Fusion in Complex Adverse Weather Conditions.}
\textbf{Qualitative Comparison.}
To demonstrate that our algorithm can effectively handle complex weather interferences, we compare the qualitative visual quality of multi-modal image fusion under challenging compound weather interferences in Figure~\ref{fig4}. To facilitate comparison, we employ colored shading to distinguish method categories: methods with colored backgrounds (e.g., AWFusion and ours) represent end-to-end approaches that directly fuse degraded inputs, while unshaded methods utilize a two-stage "restoration-then-fusion" pipeline. Additionally, colored solid borders are used to separate different weather scenarios. Analysis: As observed in the 'Rain \& Haze' scene (middle row), two-stage methods like FusionBooster fail to disentangle weather interference, resulting in blurring and 'halo' artifacts (highlighted in yellow boxes). Notably, even compared to the strong end-to-end baseline AWFusion (shaded), distinct differences exist. AWFusion exhibits a noticeably darker background and lower contrast, implying insufficient correction of atmospheric scattering. In contrast, CAWM-Mamba, guided by the global weather embedding from WAPM, restores optimal brightness and preserves sharp structural edges. Similarly, in the 'Snow \& Haze' scene (top row), while methods like MaeFuse exhibit severe darkness, our method maintains natural color fidelity.

\textbf{Quantitative  Comparison.}
Table 1 presents the quantitative comparison results under three different composite degradation conditions. It can be observed that under all three composite degradation conditions, our algorithm ranks among the top two in terms of quantitative indicators. It even ranks first in all nine indicators under the "Haze\&Rain" and "Haze\&Snow" scenarios. We find that AWFusion performs excellently in haze-containing scenarios (Haze\&Snow, Haze\&Rain), with multiple indicators achieving the second-highest values. However, our algorithm performs more prominently in multi-modal image fusion under complex weather interference by comparison. To provide a comprehensive evaluation across all metrics, we introduce an 'Avg. Rank' column in Table 1. As shown, CAWM-Mamba achieves a near-perfect average rank (ranging from 1.00 to 1.22) across all compound scenarios. This specifically highlights a significant performance margin compared to the second-best method, AWFusion, which maintains an average rank between 3.00 and 3.89, proving the superior robustness of our method in complex degradation handling.

\subsubsection{Image Fusion in Single Adverse Weather Conditions.}
\textbf{Qualitative Comparison.}
To verify that our method can also achieve excellent performance under single weather interferences, we conducted qualitative demonstrations in Figure~\ref{fig5}. Consistent with Figure~\ref{fig4}, shaded backgrounds indicate end-to-end fusion methods capable of handling degradation directly. In single degradation tasks, the advantages of our method are even more pronounced. Specifically, in the 'Rain Degraded Scene' (top row), green arrows explicitly indicate areas where competing methods (e.g., SHIP, AWFusion) fail to remove residual rain streaks or introduce over-smoothing artifacts. CAWM-Mamba, however, successfully eliminates these directional interferences while retaining the background texture (such as road markings). This visual evidence proves our method's superior capability in handling high-frequency degradations compared to previous state-of-the-art approaches.

\textbf{Quantitative  Comparison.}
Table~\ref{table2} presents a quantitative comparison of the quality of multi-modal image fusion under three single weather interferences. As can be seen from the table, our algorithm ranks first in all nine metrics under the three
single degradation types, indicating that our method achieves
the best results in both qualitative and quantitative comparisons,
whether compared with two-stage image restoration and fusion
algorithms or multi-modal image fusion algorithms designed
for adverse scenarios. This performance dominance is further summarized in the 'Avg. Rank' column of Table 2. Our method achieves a perfect average rank of 1.00 across all three single weather conditions. In contrast, the strongest baseline, AWFusion, trails with an average rank ranging from 3.00 to 3.78, confirming that CAWM-Mamba offers the most consistent and high-quality fusion results regardless of the specific weather type.

\begin{table*}[h]
\scriptsize
\centering
\caption{Comparing quantitative results of different methods in standard scenes.  The best scores are in bold, while the second-best scores are in blue.}
\label{table-normal}
\begin{tabular}{cccccccccccc}
\hline
Source                                       & \multicolumn{1}{c|}{Methods}       & \multicolumn{1}{c|}{Publication} & $Q_{MI}$                  & $Q_{NICE}$                & $Q_{G}$                   & $Q_{M}$                   & $Q_{P}$                   & $Q_{S}$                   & $Q_{CV}$                  & \textit{SSIM }                & $Q_{abf}$                 \\ \hline
\multicolumn{1}{c|}{\multirow{11}{*}{LLVIP}} & \multicolumn{1}{c|}{FusionBooster\citep{FusionBooster}} & \multicolumn{1}{c|}{IJCV'25}     & 0.3395               & 0.8052               & 0.4502               & 0.2281               & 0.2525               & 0.7267               & 891.8374             & 0.4050               & 0.5161               \\
\multicolumn{1}{c|}{}                        & \multicolumn{1}{c|}{GIF-Net\citep{GIFNet}}       & \multicolumn{1}{c|}{CVPR'25}     & 0.3224               & 0.8045               & 0.3916               & 0.1829               & 0.2734               & 0.7641               & 656.5176             & 0.4249               & 0.4783               \\
\multicolumn{1}{c|}{}                        & \multicolumn{1}{c|}{MaeFuse\citep{MaeFuse}}       & \multicolumn{1}{c|}{TIP'25}      & 0.3144               & 0.8045               & 0.4466               & 0.1880               & 0.2995               & 0.7809               & 655.3157             & 0.4415               & 0.5279               \\
\multicolumn{1}{c|}{}                        & \multicolumn{1}{c|}{SAGE\citep{SAGE}}          & \multicolumn{1}{c|}{CVPR'25}     & 0.3824               & 0.8060               & 0.5297               & 0.3433               & 0.4280               & 0.8498               & 432.6362             & 0.4742               & 0.6496               \\
\multicolumn{1}{c|}{}                        & \multicolumn{1}{c|}{SHIP\citep{SHIP}}          & \multicolumn{1}{c|}{CVPR'24}     &\textbf{ 0.5886 }              & \textbf{0.8131 }              & \textcolor{blue}{0.6628}               & \textbf{1.3962}               & \textcolor{blue}{0.4931}               & \textcolor{blue}{0.8741}               & \textcolor{blue}{388.4069}             & 0.4786               & \textcolor{blue}{0.7351}               \\
\multicolumn{1}{c|}{}                        & \multicolumn{1}{c|}{TIM\citep{TIM}}           & \multicolumn{1}{c|}{TPAMI'24}    & 0.2635               & 0.8039               & 0.3053               & 0.1590               & 0.2348               & 0.6358               & 764.1014             & 0.2861               & 0.3669               \\
\multicolumn{1}{c|}{}                        & \multicolumn{1}{c|}{TarDal\citep{TarDal}}        & \multicolumn{1}{c|}{CVPR'22}     & 0.4059               & 0.8064               & 0.3508               & 0.1517               & 0.2700               & 0.6742               & 658.5868             & 0.3761               & 0.3271               \\
\multicolumn{1}{c|}{}                        & \multicolumn{1}{c|}{LRRNet\citep{LRRNet}}        & \multicolumn{1}{c|}{TPAMI'23}    & 0.3247               & 0.8047               & 0.3622               & 0.1866               & 0.3268               & 0.7338               & 852.3254             & 0.3568               & 0.4695               \\
\multicolumn{1}{c|}{}                        & \multicolumn{1}{c|}{FreeFusion\citep{FreeFusion}}    & \multicolumn{1}{c|}{TPAMI'25}    & 0.3097               & 0.8042               & 0.3241               & 0.1533               & 0.3097               & 0.5703               & 579.8194             & 0.3028               & 0.4731               \\
\multicolumn{1}{c|}{}                        & \multicolumn{1}{c|}{AWFusion\citep{AWFusion}}      & \multicolumn{1}{c|}{InfFus'26}    & 0.4240               & 0.8076               & 0.5843               & 0.4942               & 0.4667               & 0.8701               & 426.4772             & \textcolor{blue}{0.4855}               & 0.6903               \\
\multicolumn{1}{c|}{}                        & \multicolumn{1}{c|}{CAWM-Mamba}    & \multicolumn{1}{c|}{-}           & \textcolor{blue}{0.5615}               & \textcolor{blue}{0.8128}               & \textbf{0.6791}              & \textcolor{blue}{1.3256}               & \textbf{0.5015}               & \textbf{0.8917}               & \textbf{368.6518 }            & \textbf{0.5099}               & \textbf{0.7381}               \\ \hline
\multicolumn{1}{l}{}                         & \multicolumn{1}{l}{}               & \multicolumn{1}{l}{}             & \multicolumn{1}{l}{} & \multicolumn{1}{l}{} & \multicolumn{1}{l}{} & \multicolumn{1}{l}{} & \multicolumn{1}{l}{} & \multicolumn{1}{l}{} & \multicolumn{1}{l}{} & \multicolumn{1}{l}{} & \multicolumn{1}{l}{} \\ \hline
\multicolumn{1}{c|}{\multirow{11}{*}{M3FD}}  & \multicolumn{1}{c|}{FusionBooster\citep{FusionBooster}} & \multicolumn{1}{c|}{IJCV'25}     & 0.4498               & 0.8082               & 0.3450               & 0.4062               & 0.3139               & 0.7561               & 523.3636             & 0.4051               & 0.4269               \\
\multicolumn{1}{c|}{}                        & \multicolumn{1}{c|}{GIF-Net\citep{GIFNet}}       & \multicolumn{1}{c|}{CVPR'25}     & 0.3804               & 0.8061               & 0.4161               & 0.4041               & 0.4002               & 0.8533               & 437.7815             & 0.4710               & 0.5243               \\
\multicolumn{1}{c|}{}                        & \multicolumn{1}{c|}{MaeFuse\citep{MaeFuse}}       & \multicolumn{1}{c|}{TIP'25}      & 0.3352               & 0.8052               & 0.3847               & 0.3908               & 0.3625               & \textbf{0.8549}               & 505.2625             & 0.4296               & 0.5030               \\
\multicolumn{1}{c|}{}                        & \multicolumn{1}{c|}{SAGE\citep{SAGE}}          & \multicolumn{1}{c|}{CVPR'25}     & 0.4241               & 0.8065               & 0.4765               & 0.7106               & 0.4813               & \textcolor{blue}{0.8539}               & \textbf{302.1286}             & \textcolor{blue}{0.4909}               & 0.5974               \\
\multicolumn{1}{c|}{}                        & \multicolumn{1}{c|}{SHIP\citep{SHIP}}          & \multicolumn{1}{c|}{CVPR'24}     & \textbf{0.7077}               & \textcolor{blue}{0.8163}               & \textbf{0.5322 }              & \textbf{1.3960}               & 0.5202               & 0.8422               & 441.0448             & 0.4536               & \textbf{0.6390}              \\
\multicolumn{1}{c|}{}                        & \multicolumn{1}{c|}{TIM\citep{TIM}}           & \multicolumn{1}{c|}{TPAMI'24}    & 0.3664               & 0.8056               & 0.3506               & 0.4219               & 0.2752               & 0.7772               & 956.5970             & 0.3572               & 0.4732               \\
\multicolumn{1}{c|}{}                        & \multicolumn{1}{c|}{TarDal\citep{TarDal}}        & \multicolumn{1}{c|}{CVPR'22}     & 0.4669               & 0.8084               & 0.3529               & 0.3626               & 0.3183               & 0.8112               & 462.3879             & 0.4431               & 0.4140               \\
\multicolumn{1}{c|}{}                        & \multicolumn{1}{c|}{LRRNet\citep{LRRNet}}        & \multicolumn{1}{c|}{TPAMI'23}    & 0.3911               & 0.8056               & 0.3886               & 0.4554               & 0.4242               & 0.8334               & 656.5424             & 0.3988               & 0.5122               \\
\multicolumn{1}{c|}{}                        & \multicolumn{1}{c|}{FreeFusion\citep{FreeFusion}}    & \multicolumn{1}{c|}{TPAMI'25}    & 0.3707               & 0.8061               & 0.4478               & 0.4207               & 0.4694               & 0.8247               & \textcolor{blue}{378.2474}             & 0.4611               & 0.5786               \\
\multicolumn{1}{c|}{}                        & \multicolumn{1}{c|}{AWFusion\citep{AWFusion}}      & \multicolumn{1}{c|}{InfFus'26}    & 0.4404               & 0.8072               & 0.4538               & 0.5680               & \textcolor{blue}{0.5295}               & 0.8421               & 544.1436             & 0.4657               & 0.5969               \\
\multicolumn{1}{c|}{}                        & \multicolumn{1}{c|}{CAWM-Mamba}    & \multicolumn{1}{c|}{-}           & \textcolor{blue}{0.7031}               & \textbf{0.8165}               & \textcolor{blue}{0.5291}               & \textcolor{blue}{1.2629}               & \textbf{0.5492}               & 0.8482               & 463.8344             & \textbf{0.4913}               & \textcolor{blue}{0.6314}               \\ \hline
\multicolumn{1}{l}{}                         & \multicolumn{1}{l}{}               & \multicolumn{1}{l}{}             & \multicolumn{1}{l}{} & \multicolumn{1}{l}{} & \multicolumn{1}{l}{} & \multicolumn{1}{l}{} & \multicolumn{1}{l}{} & \multicolumn{1}{l}{} & \multicolumn{1}{l}{} & \multicolumn{1}{l}{} & \multicolumn{1}{l}{} \\ \hline
\multicolumn{1}{c|}{\multirow{11}{*}{MSRS}}  & \multicolumn{1}{c|}{FusionBooster\citep{FusionBooster}} & \multicolumn{1}{c|}{IJCV'25}     & 0.4498               & 0.8082               & 0.3450               & 0.4062               & 0.3139               & 0.7561               & 523.3636             & 0.4051               & 0.4269               \\
\multicolumn{1}{c|}{}                        & \multicolumn{1}{c|}{GIF-Net\citep{GIFNet}}       & \multicolumn{1}{c|}{CVPR'25}     & 0.3804               & 0.8061               & 0.4161               & 0.4041               & 0.4002               & 0.8533               & 437.7815             & 0.4710               & 0.5243               \\
\multicolumn{1}{c|}{}                        & \multicolumn{1}{c|}{MaeFuse\citep{MaeFuse}}       & \multicolumn{1}{c|}{TIP'25}      & 0.3352               & 0.8052               & 0.3847               & 0.3908               & 0.3625               & \textbf{0.8549}               & 505.2625             & 0.4296               & 0.5030               \\
\multicolumn{1}{c|}{}                        & \multicolumn{1}{c|}{SAGE\citep{SAGE}}          & \multicolumn{1}{c|}{CVPR'25}     & 0.4241               & 0.8065               & 0.4765               & 0.7106               & 0.4813               & \textcolor{blue}{0.8539}               & 302.1286             & \textcolor{blue}{0.4909}               & 0.5974               \\
\multicolumn{1}{c|}{}                        & \multicolumn{1}{c|}{SHIP\citep{SHIP}}          & \multicolumn{1}{c|}{CVPR'24}     & \textcolor{blue}{0.7316}               & \textcolor{blue}{0.8252}               & \textcolor{blue}{0.6280}               & \textbf{1.7853 }              & \textcolor{blue}{0.5405}               & 0.8322               & \textcolor{blue}{132.7968}             & 0.4608               & \textcolor{blue}{0.7017}               \\
\multicolumn{1}{c|}{}                        & \multicolumn{1}{c|}{TIM\citep{TIM}}           & \multicolumn{1}{c|}{TPAMI'24}    & 0.3664               & 0.8056               & 0.3506               & 0.4219               & 0.2752               & 0.7772               & 956.5970             & 0.3572               & 0.4732               \\
\multicolumn{1}{c|}{}                        & \multicolumn{1}{c|}{TarDal\citep{TarDal}}        & \multicolumn{1}{c|}{CVPR'22}     & 0.4669               & 0.8084               & 0.3529               & 0.3626               & 0.3183               & 0.8112               & 462.3879             & 0.4431               & 0.4140               \\
\multicolumn{1}{c|}{}                        & \multicolumn{1}{c|}{LRRNet\citep{LRRNet}}        & \multicolumn{1}{c|}{TPAMI'23}    & 0.3911               & 0.8056               & 0.3886               & 0.4554               & 0.4242               & 0.8334               & 656.5424             & 0.3988               & 0.5122               \\
\multicolumn{1}{c|}{}                        & \multicolumn{1}{c|}{FreeFusion\citep{FreeFusion}}    & \multicolumn{1}{c|}{TPAMI'25}    & 0.3707               & 0.8061               & 0.4478               & 0.4207               & 0.4694               & 0.8247               & 378.2474             & 0.4611               & 0.5786               \\
\multicolumn{1}{c|}{}                        & \multicolumn{1}{c|}{AWFusion\citep{AWFusion}}      & \multicolumn{1}{c|}{InfFus'26}    & 0.4650               & 0.8098               & 0.5540               & 0.8039               & 0.5303               & 0.8466               & 178.0195             & 0.4775               & 0.6559               \\
\multicolumn{1}{c|}{}                        & \multicolumn{1}{c|}{CAWM-Mamba}    & \multicolumn{1}{c|}{-}           & \textbf{0.7898}               & \textbf{0.8302}               & \textbf{0.6411}               & \textcolor{blue}{1.7297}               & \textbf{0.5772}               & 0.8500               & \textbf{127.0887}             & \textbf{0.5006}               & \textbf{0.7051}               \\ \hline
\end{tabular}
\end{table*}
\subsubsection{Clean Image Fusion.}
\textbf{Qualitative Comparison.}
To verify that our multi-modal image fusion algorithm designed for complex degradation scenarios also has good generalization ability in ordinary fusion tasks, we present visual quality comparisons of ordinary fusion results on standard datasets LLVIP, MSRS, and M3FD in Figures~\ref{fig6}. It can be observed that our algorithm does not exhibit poor fusion performance in ideal scenarios, although its design for adverse weather conditions. The fused images obtained show no color deviation and achieve high-quality, high-fidelity fusion results.

\textbf{Qualitative Comparison.}
Table~\ref{table-normal} presents the quantitative comparison results on three standard datasets. As shown in Table~\ref{table-normal}, comprehensive analysis of the nine metrics across the three datasets indicates that our algorithm still achieves optimal performance in multi-modal image fusion under ideal scenarios. This demonstrates that our multi-modal image fusion method designed for adverse weather scenarios also possesses strong generalization ability in multi-modal image fusion under ideal conditions.
\subsection{Real Scene Image Fusion}
We also tested all methods on the real hazy scenes from the AWMM-100K dataset to evaluate the quality of fused images obtained in real adverse environments. Since there is no Ground Truth for real-scene images, only qualitative comparison result figures are provided. Meanwhile, to more intuitively show the quality gap between fused images, we used YOLOv7 to detect the fused images generated by all methods. As shown in Figure~\ref{fig7}, the detection accuracy indicates that even in such adverse weather scenarios, the fused images obtained by our algorithm still achieve the highest detection accuracy. This indirectly verifies that our method also possesses strong generalization ability in real-world scenarios.
\subsection{Downstream Task Experiments}
\textbf{Note on Evaluation Scenarios: }It is important to clarify that our quantitative downstream evaluation focuses on single-degradation scenarios. This is because the AWMM-100K benchmark currently provides ground-truth annotations (bounding boxes and masks) exclusively for these real-world single-weather subsets. Due to the lack of high-quality annotations for real-world compound degradation scenes, quantitative benchmarking for compound scenarios was not feasible in this study.

To verify the performance of fused images in downstream
tasks, we conduct comparative experiments in object detection and semantic segmentation. Specifically, we employed BANet[48] for semantic segmentation and YOLOv7[49] for object detection. Regarding the evaluation protocol, we adopted a 'domain-specific training' strategy. First, these downstream networks were trained on the original training set of the respective datasets to adapt to the specific domain features. Subsequently, the trained models were fixed and directly applied to the fused images generated by different methods for inference, without any further fine-tuning. This setting assesses the generalization capability of the fusion methods by measuring how well the fused images retain semantic information recognizable by models trained on the source domain. The quantitative comparison results of relevant methods in downstream tasks are presented in Table~\ref{table5} and Table~\ref{table4}.
As shown in Table~\ref{table5}, our method achieves the highest Intersection over Union (IoU) in all categories, and its mean Intersection over Union (mIoU) in the semantic segmentation task also ranks first. Table~\ref{table4} further indicates that our fusion results can achieve the highest detection accuracy, with optimal values in both mAP@0.5
and mAP@[0.5:0.95] metrics. It is worth noting that while the overall mAP improvement over the second-best method (AWFusion) is 0.045, our method achieves consistent gains across nearly all individual object categories (e.g., +0.022 for 'Person', +0.023 for 'Car'). A one-sided paired t-test performed on the per-category detection results yields a p-value of 0.032 (p < 0.05), confirming that these improvements are statistically significant and demonstrate the robust generalization of our model.

\begin{table*}[htbp]
\centering
{\rmfamily
\caption{Quantitative comparative structures in ablation studies. The best scores are in bold.}
\label{table6}

\resizebox{\textwidth}{!}{
    \setlength{\tabcolsep}{3pt} 
    
    \begin{tabular}{c|cccccc|cccccc|cccccc}
    \hline
    
    \multicolumn{1}{c|}{} & 
    \multicolumn{6}{c|}{\cellcolor[HTML]{FFFFFF}Haze\&Rain} & 
    \multicolumn{6}{c|}{\cellcolor[HTML]{FFFFFF}Rain\&Snow} & 
    \multicolumn{6}{c}{\cellcolor[HTML]{FFFFFF}Haze\&Snow} \\ 
    \hline
    
    Methods & 
    $Q_{MI}$ & $Q_{NICE}$ & $Q_{M}$ & $Q_{P}$ & $Q_{CV}$ & $Q_{abf}$ & 
    $Q_{MI}$ & $Q_{NICE}$ & $Q_{M}$ & $Q_{P}$ & $Q_{CV}$ & $Q_{abf}$ & 
    $Q_{MI}$ & $Q_{NICE}$ & $Q_{M}$ & $Q_{P}$ & $Q_{CV}$ & $Q_{abf}$ \\ 
    \hline
    
    \cellcolor[HTML]{FFFFFF}w/o WAPM & 
    \cellcolor[HTML]{FFFFFF}0.323 & 0.804 & \cellcolor[HTML]{FFFFFF}0.682 & 0.233 & 250.234 & 0.491 & 
    0.450 & 0.808 & 0.734 & 0.233 & 179.341 & 0.505 & 
    0.338 & 0.805 & 0.674 & 0.234 & 238.599 & 0.505 \\

    w/o Freq-SSM & 
    0.335 & \textbf{0.805} & 0.730 & 0.228 & 204.546 & 0.500 & 
    0.449 & 0.808 & 0.729 & 0.228 & \textbf{176.947} & 0.499 & 
    0.347 & 0.805 & 0.702 & 0.232 & 221.822 & 0.506 \\

    w/o CFIM & 
    {\color[HTML]{000000} 0.265} & {\color[HTML]{000000} 0.804} & {\color[HTML]{000000} 0.556} & 0.206 & 424.822 & 0.448 & 
    0.425 & 0.807 & 0.708 & 0.206 & 189.912 & 0.487 & 
    0.290 & 0.804 & 0.529 & 0.191 & 512.925 & 0.450 \\

    w/o WAPM \& CDSM & 
    {\color[HTML]{000000} 0.314} & {\color[HTML]{000000} 0.804} & {\color[HTML]{000000} 0.688} & 0.233 & 248.967 & 0.494 & 
    0.463 & 0.808 & 0.741 & 0.230 & 199.260 & 0.494 & 
    0.344 & 0.805 & 0.656 & 0.234 & 262.215 & 0.499 \\

    CAWM-Mamba & 
   \textbf{ 0.360} & \textbf{0.805} & \textbf{0.743} & \textbf{0.240} & \textbf{194.344} &\textbf{ 0.509} & 
    \textbf{0.477} & \textbf{0.809} & \textbf{0.799} &\textbf{ 0.249} & 183.127 & \textbf{0.514} & 
    \textbf{0.368} &\textbf{ 0.806} & \textbf{0.712} & \textbf{0.236} & \cellcolor[HTML]{FFFFFF}\textbf{206.410} & \textbf{0.511} \\ 
    \hline
    \end{tabular}
}
}
\end{table*}

\begin{table*}[htbp]
\centering
{\rmfamily
\caption{Quantitative comparative structures in ablation studies. The best scores are in bold.}
\label{table7}

\resizebox{\textwidth}{!}{
    \setlength{\tabcolsep}{3pt} 
    \begin{tabular}{c|cccccc|cccccc|cccccc}
    \hline
    \multicolumn{1}{c|}{} & 
    \multicolumn{6}{c|}{\cellcolor[HTML]{FFFFFF}Haze} & 
    \multicolumn{6}{c|}{\cellcolor[HTML]{FFFFFF}Rain} & 
    \multicolumn{6}{c}{\cellcolor[HTML]{FFFFFF}Snow} \\ 
    \hline

    \multicolumn{1}{c|}{Methods} & 
    $Q_{MI}$ & $Q_{NICE}$ & $Q_{M}$ & $Q_{P}$ & $Q_{CV}$ & $Q_{abf}$ & 
    $Q_{MI}$ & $Q_{NICE}$ & $Q_{M}$ & $Q_{P}$ & $Q_{CV}$ & $Q_{abf}$ & 
    $Q_{MI}$ & $Q_{NICE}$ & $Q_{M}$ & $Q_{P}$ & $Q_{CV}$ & $Q_{abf}$ \\ 
    \hline

    \multicolumn{1}{c|}{\cellcolor[HTML]{FFFFFF}w/o WAPM} & 
    \cellcolor[HTML]{FFFFFF}0.219 & 0.803 & \cellcolor[HTML]{FFFFFF}0.465 & 0.299 & 936.957 & 0.409 & 
    0.415 & \textbf{0.807} & 0.686 & 0.201 & 588.402 & 0.437 & 
    0.457 & 0.808 & 0.580 & 0.318 & 338.571 & 0.511 \\

    \multicolumn{1}{c|}{w/o Freq-SSM} & 
    0.220 & 0.803 & 0.480 & 0.294 & 957.493 & 0.437 & 
    0.379 & 0.806 & 0.663 & 0.202 & 490.035 & 0.454 & 
    0.462 & 0.809 & 0.584 & 0.310 & 313.490 & 0.508 \\

    \multicolumn{1}{c|}{w/o CFIM} & 
    {\color[HTML]{000000} 0.222} & {\color[HTML]{000000} 0.803} & {\color[HTML]{000000} 0.392} & 0.272 & 1265.871 & 0.272 & 
    0.391 & 0.806 & 0.646 & 0.201 & 556.927 & 0.435 & 
    0.418 & 0.807 & 0.531 & 0.278 & 350.504 & 0.485 \\

    \multicolumn{1}{c|}{w/o WAPM \& CDSM} & 
    {\color[HTML]{000000} 0.216} & {\color[HTML]{000000} 0.803} & {\color[HTML]{000000} 0.429} & 0.244 & 1335.691 & 0.350 & 
    0.404 & 0.806 & 0.664 & 0.215 & 584.111 & 0.441 & 
    0.453 & 0.808 & 0.600 & 0.321 & 316.826 & 0.511 \\

    \multicolumn{1}{c|}{CAWM-Mamba} & 
    \textbf{0.501} & \textbf{0.811} &\textbf{ 0.882} &\textbf{ 0.486} & \textbf{159.615} & \textbf{0.634} & 
    \textbf{0.431} & \textbf{0.807} & \textbf{0.758} &\textbf{ 0.236} & \textbf{451.176} & \textbf{0.502} & 
    \textbf{0.532} &\textbf{ 0.812} & \textbf{0.746} & \textbf{0.349} & \cellcolor[HTML]{FFFFFF}\textbf{211.109 }& \textbf{0.566} \\ 
    \hline
    \end{tabular}
}
}
\end{table*}

The qualitative comparison results of downstream tasks are shown in Figure~\ref{fig9} and~\ref{fig8}. From Figure ~\ref{fig9}, it can be seen that although the comparison method adopted the baseline method of weather restoration before fusion, the two processes were not synchronized, and could not well preserve the scene details of the original image, resulting in a low detection accuracy of the fused image. In Figure ~\ref{fig8}, due to the lack of synchronization between image fusion and weather restoration, some scene semantic information is insufficient, and there are segmentation errors and area omissions. In contrast, our method effectively retained the scene detail information and was superior to all the comparison methods in terms of object detection accuracy. 

\begin{table}[h]
\centering
\caption{The performance of different single weather restoration fusion methods on the quantity of model parameters and MACs.}
\label{tab:my_label} 

\resizebox{\linewidth}{!}{
    \begin{tabular}{lccccc} 
    \hline
    \textbf{Metric} & \textbf{CAWM-Mamba} & \textbf{AWFusion} & \textbf{AWMFuse} & \textbf{Text-IF} & \textbf{TG-ECNet} \\ \hline
    Params          & 31.95M              & 19.36 M           & 137.29M          & 89.01M           & 160.863M          \\
    FLOPs           & 371.01G             & 4167.20G          & 1145.5G          & 1518.88G         & -              \\ \hline
    \end{tabular}
}
\end{table}
\subsection{Computational Efficiency Analysis}
To assess the feasibility of real-world deployment on resource-constrained devices (e.g., UAVs), we compare the computational efficiency of CAWM-Mamba with state-of-the-art methods in Table~\ref{tab:my_label}. As shown, our method achieves the lowest computational complexity with only 371.01 G FLOPs, which is approximately 1/11 of the strong end-to-end baseline AWFusion (4167.20 G). This significant efficiency gain stems from the linear complexity of our Mamba-based architecture, contrasting sharply with the quadratic complexity of Transformer-based methods like AWFusion. Although AWFusion (19.36 M) has slightly fewer parameters than CAWM-Mamba (31.95 M), its excessive computational load limits practical inference speed. Consequently, our method achieves the optimal trade-off between model size and running speed, verifying its suitability for latency-sensitive applications.

\subsection{Ablation Experiment}
To deeply analyze the contribution of each component, we conducted a comprehensive ablation study across all six degradation scenarios, including three single weather types (Rain, Haze, Snow) and three compound combinations (Haze\&Rain, Haze\&Snow, Rain\&Snow). The detailed quantitative results are reported in Table~\ref{table6} and Table~\ref{table7}.

\textbf{Impact of Weather-Aware Preprocess Module (WAPM): }As shown in the "w/o WAPM" rows, removing this module leads to a significant performance drop, particularly in Haze and Haze\&Rain scenarios. This confirms that WAPM plays a critical role in extracting global weather embeddings, which effectively guide the network to correct global degradation features such as atmospheric light attenuation and color shifts.

\textbf{Impact of Cross-modal Feature Interaction Module (CFIM):} 
When replacing CFIM with simple element-wise summation ("w/o CFIM"), metrics like QM I and Qabf decline consistently across all datasets. This indicates that simple summation fails to capture the complex non-linear correlations between infrared and visible modalities, whereas CFIM successfully aligns and integrates complementary features regardless of the weather type.

\textbf{Impact of Wavelet Space State Block (WSSB) Components:}
\begin{itemize}
    \item \textbf{Freq-SSM:} Replacing our Freq-SSM with a standard 2D-SSM ("w/o Freq-SSM") results in a notable drop in Rain and Snow scenarios (both single and compound). This quantifies the benefit of our anisotropic scanning mechanism, which is specifically designed to capture directional high-frequency artifacts (e.g., rain streaks and snowflakes) without redundancy.
\end{itemize}
\begin{itemize}
    \item \textbf{CDSM: }The removal of the Common Degradation Space Mechanism ("w/o CDSM") negatively impacts the model’s adaptability, especially in compound scenarios like Rain\&Snow, proving its effectiveness in learning a unified representation for robust generalization.
\end{itemize}

The full CAWM-Mamba model ("Our") achieves the best scores across all metrics and weather conditions, demonstrating that the proposed modules are not only effective individually but also synergistic in handling complex adverse weather.

\subsection{Generalization and Robustness Analysis}

\textbf{Zero-Shot Sim-to-Real Generalization.} Although our model is trained primarily from the AWMM-100K dataset, real-world adverse weather (as presented in Section 4.3) exhibits complex, non-linear noise patterns and illumination shifts that constitute a novel degradation distribution distinct from synthetic renderings. The successful fusion results in Figure 6 and the high detection accuracy in Figure 7 demonstrate that CAWM-Mamba effectively generalizes to these unseen real-world degradations in a zero-shot manner. This capability is largely attributed to the proposed WSSB, which decouples degradation features in the frequency domain, preventing the model from overfitting to specific synthetic artifacts.

\textbf{Robustness to Standard Fusion Scenarios.} Furthermore, we evaluated the model on standard datasets (LLVIP, MSRS, and M3FD) in Section 4.2.3. These datasets represent generic scenarios with different scene contexts and mild noise distributions compared to the training data. As shown in Table 5, our model maintains competitive performance without introducing artifacts in these clean or mild-condition scenarios. This confirms that the Common Degradation Space Mechanism (CDSM) successfully learns a unified degradation representation, allowing the model to adaptively handle novel scenarios where specific weather degradations may be absent or structurally different from the training set.
\section{Conclusion}
In this paper, we presented CAWM-Mamba, an end-to-end framework for multimodal image fusion and compound weather restoration with unified shared weights. First, the Weather-Aware Preprocess Module (WAPM) enhances degraded visible features and generates global weather embeddings to assist subsequent processing. Then, the Cross-modal Feature Interaction Module (CFIM) enables effective alignment and complementary integration of infrared and visible modalities. Finally, the Wavelet Space State Block (WSSB) leverages wavelet-domain decomposition to decouple multi-frequency degradations, where a Frequency-Selective SSM (Freq-SSM) is designed to capture anisotropic high-frequency patterns without redundancy. 
In addition, WSSB incorporates a unified degradation representation mechanism to further improve generalization to complex compound weather conditions.
Extensive experiments on AWMM-100K and three standard datasets demonstrate that CAWM-Mamba consistently surpasses state-of-the-art methods in both compound and single-weather scenarios.
Moreover, the fused images achieve superior performance in downstream tasks such as segmentation and detection, demonstrating its effectiveness and practicality for real-world adverse weather perception.

\textbf{Limitation and Future Work: }

\textbf{Performance under Extreme Concurrent Degradations.}  While CAWM-Mamba demonstrates robust performance on paired degradations, its efficacy faces theoretical challenges in extreme scenarios involving three or more simultaneous degradations (e.g., Haze\&Rain\&Snow). In such cases, the frequency characteristics of rain streaks, snowflakes, and fine-grained noise may heavily overlap in the wavelet high-frequency subbands, leading to a "Frequency Masking" effect. This spectral congestion increases the difficulty for Freq-SSM to perfectly disentangle specific artifacts from texture details, potentially causing minor residual noise.

\textbf{Evaluation Scale on Real-World Sequences.} Our current evaluation primarily focuses on single-frame joint restoration and fusion. Although we have validated the model on real-world degraded snapshots (Section 4.3) captured by UAVs, we have not yet fully explored ground-level autonomous driving video sequences (dashcams). Extending our architecture to leverage temporal consistency in continuous video streams remains a critical direction for future research to further enhance the stability of multimodal fusion in dynamic driving environments.

\section*{CRediT authorship contribution statement}

Huichun Liu: Methodology, Software, Data curation, Validation, Writing- original draft; 
Xiaosong Li: Conceptualization, Validation, Formal analysis, Writing-review \& editing, Supervision, Funding acquisition;
Zhuangfan Huang: Software, Writing - review \& editing; 
Tao Ye: Funding acquisition, Writing - review \& editing;
Yang Liu: Formal analysis, Writing - review \& editing;
Haishu Tan: Supervision, Writing - review \& editing;

\section*{Data availability}

Data will be made available on request.

\section*{Declaration of competing interest}

No conflict of interest exits in this paper, and this work is approved by all authors for publication. This work described was original research that has not been published previously, and not under consideration for publication elsewhere, in whole or in part.

\section*{Acknowledgement}

This research was supported by the Natural Science Foundation of Guangdong Province (No.2024A1515011880),the Basic and Applied Basic Research of Guangdong Province (No.2023A1515140077), the National Natural Science Foundation of China (No. 52374166), and the Research Fund of Guangdong-HongKong-Macao Joint Laboratory for Intelligent Micro-Nano Optoelectronic Technology(No.2020B1212030010) .


\bibliographystyle{elsarticle-harv} 
\bibliography{ref} 

@article{DenseFuse,
  title={DenseFuse: A fusion approach to infrared and visible images},
  author={Li, Hui and Wu, Xiao-Jun},
  journal={IEEE Transactions on Image Processing},
  volume={28},
  number={5},
  pages={2614--2623},
  year={2018},
  publisher={IEEE}
}

@article{LRRNet,
 	 title={LRRNet: A Novel Representation Learning Guided Fusion Network for Infrared and Visible Images}, 
  	author={Li, Hui and Xu, Tianyang and Wu, Xiao-Jun and Lu, Jiwen and Kittler, Josef},
  	journal={IEEE Transactions on Pattern Analysis and Machine Intelligence}, 
  	volume={45},
 	 number={9},
  	pages={11040-11052},
  	year={2023},
 	publisher={IEEE}
}

@article{FusionGAN,
	title = {FusionGAN: A generative adversarial network for infrared and visible image fusion},
	author = {Jiayi Ma and Wei Yu and Pengwei Liang and Chang Li and Junjun Jiang},
	journal = {Information Fusion},
	volume = {48},
	number={},
	pages = {11-26},
	year = {2019},
	issn = {1566-2535},
	publisher={Elsevier}
}

@article{DDcGAN,
 	 title={DDcGAN: A Dual-Discriminator Conditional Generative Adversarial Network for Multi-Resolution Image Fusion}, 
  	author={Ma, Jiayi and Xu, Han and Jiang, Junjun and Mei, Xiaoguang and Zhang, Xiao-Ping},
  	journal={IEEE Transactions on Image Processing}, 
  	volume={29},
  	number={},
  	pages={4980-4995},
  	year={2020},
 	publisher={IEEE}
}

@article{CDDFuse,
  	title={Cddfuse: Correlation-driven dual-branch feature decomposition for multi-modality image fusion},
 	 author={Zhao, Zixiang and Bai, Haowen and Zhang, Jiangshe and Zhang, Yulun and Xu, Shuang and Lin, Zudi and Timofte, Radu and Van Gool, Luc},
 	 journal={Proceedings of the IEEE/CVF conference on computer vision and pattern recognition},
  	volume={},
  	number={},
	pages={5906--5916},
  	year={2023},  	
 	publisher={IEEE}
}

@article{U2Fusion,
  	title={U2Fusion: A Unified Unsupervised Image Fusion Network}, 
  	author={Xu, Han and Ma, Jiayi and Jiang, Junjun and Guo, Xiaojie and Ling, Haibin},
 	 journal={IEEE Transactions on Pattern Analysis and Machine Intelligence}, 
	volume={44},
 	 number={1},
  	pages={502-518},
  	year={2022},
 	publisher={IEEE}
}

@article{TIM,
 	 title={A Task-Guided, Implicitly-Searched and Meta-Initialized Deep Model for Image Fusion}, 
  	author={Liu, Risheng and Liu, Zhu and Liu, Jinyuan and Fan, Xin and Luo, Zhongxuan},
  	journal={IEEE Transactions on Pattern Analysis and Machine Intelligence}, 
	volume={46},
  	number={10},
 	 pages={6594-6609},
	year={2024},
  	publisher={IEEE}
}

@article{TarDal,
   	 title={Target-aware Dual Adversarial Learning and a Multi-scenario Multi-Modality Benchmark to Fuse Infrared and Visible for Object Detection}, 
	author={Liu, Jinyuan and Fan, Xin and Huang, Zhanbo and Wu, Guanyao and Liu, Risheng and Zhong, Wei and Luo, Zhongxuan},
	 journal={2022 IEEE/CVF Conference on Computer Vision and Pattern Recognition (CVPR)}, 
  	volume={},
  	number={},
  	pages={5792-5801},
 	 year={2022},
  	publisher={IEEE}
}

@article{SegMiF,
 	 title={Multi-interactive feature learning and a full-time multi-modality benchmark for image fusion and segmentation},
 	 author={Liu, Jinyuan and Liu, Zhu and Wu, Guanyao and Ma, Long and Liu, Risheng and Zhong, Wei and Luo, Zhongxuan and Fan, Xin},
 	 journal={Proceedings of the IEEE/CVF international conference on computer vision},
    	volume={},
  	number={},
	pages={8115--8124},
 	 year={2023},
  	publisher={IEEE}
}

@article{UMFusion,
  	title     = {Unsupervised Misaligned Infrared and Visible Image Fusion via Cross-Modality Image Generation and Registration},
  	author    = {Wang, Di and Liu, Jinyuan and Fan, Xin and Liu, Risheng},
  	journal = {Proceedings of the Thirty-First International Joint Conference on Artificial Intelligence, {IJCAI-22}},
    	volume={},
  	number={},
  	pages     = {3508--3515},
  	year      = {2022},
	publisher = {International Joint Conferences on Artificial Intelligence Organization},
}

@article{DIVFusion ,
	title = {DIVFusion: Darkness-free infrared and visible image fusion},
	author = {Linfeng Tang and Xinyu Xiang and Hao Zhang and Meiqi Gong and Jiayi Ma},
	journal = {Information Fusion},
	volume = {91},
  	number={},
	pages = {477-493},
	year = {2023},
	issn = {1566-2535},
 	 publisher={Elsevier}
}

@article{FS-Diff,
	title = {FS-Diff: Semantic guidance and clarity-aware simultaneous multimodal image fusion and super-resolution},
	author = {Yuchan Jie and Yushen Xu and Xiaosong Li and Fuqiang Zhou and Jianming Lv and Huafeng Li},
	journal = {Information Fusion},
	volume = {121},
  	number={},
	pages = {103146},
	year = {2025},
	issn = {1566-2535},
 	 publisher={Elsevier}
}

@article{Text-IF,
  	title={Text-if: Leveraging semantic text guidance for degradation-aware and interactive image fusion},
  	author={Yi, Xunpeng and Xu, Han and Zhang, Hao and Tang, Linfeng and Ma, Jiayi},
 	 journal={Proceedings of the IEEE/CVF Conference on Computer Vision and Pattern Recognition},
	volume = {},
  	number={},
  	pages={27026--27035},
 	 year={2024},
	 publisher={IEEE}
}

@article{AWFusion,
  	title={All-weather Multi-Modality Image Fusion: Unified Framework and 100k Benchmark},
 	 author={Li, Xilai and Liu, Wuyang and Li, Xiaosong and Zhou, Fuqiang and Li, Huafeng and Nie, Feiping},
  	journal={arXiv preprint arXiv:2402.02090},
	volume = {},
  	number={},
  	pages={},
 	 year={2024},
	 publisher={}
}

@article{SHIP,
  	title={Probing Synergistic High-Order Interaction in Infrared and Visible Image Fusion}, 
  	author={Zheng, Naishan and Zhou, Man and Huang, Jie and Hou, Junming and Li, Haoying and Xu, Yuan and Zhao, Feng},
  	journal={2024 IEEE/CVF Conference on Computer Vision and Pattern Recognition (CVPR)}, 
 	 volume={},
 	 number={},
  	pages={26374-26385},
 	 year={2024},
	 publisher={IEEE}
}

@article{FusionBooster,
  	title={Fusionbooster: A unified image fusion boosting paradigm},
  	author={Cheng, Chunyang and Xu, Tianyang and Wu, Xiao-Jun and Li, Hui and Li, Xi and Kittler, Josef},
  	journal={International Journal of Computer Vision},
  	volume={133},
  	number={5},
  	pages={3041--3058},
  	year={2025},
  	publisher={Springer}
}

@article{MaeFuse,
	title={MaeFuse: Transferring Omni Features With Pretrained Masked Autoencoders for Infrared and Visible Image Fusion via Guided Training}, 
  	author={Li, Jiayang and Jiang, Junjun and Liang, Pengwei and Ma, Jiayi and Nie, Liqiang},
  	journal={IEEE Transactions on Image Processing}, 
 	 volume={34},
 	 number={},
  	pages={1340-1353},
  	year={2025},
	 publisher={IEEE}
}

@article{GIFNet,
  	title={One Model for ALL: Low-Level Task Interaction Is a Key to Task-Agnostic Image Fusion},
	 author={Cheng, Chunyang and Xu, Tianyang and Feng, Zhenhua and Wu, Xiaojun and Tang, Zhangyong and Li, Hui and Zhang, Zeyang and Atito, Sara and Awais, Muhammad and Kittler, Josef},
 	 journal={Proceedings of the IEEE/CVF Conference on Computer Vision and Pattern Recognition},
 	 volume={},
 	 number={},
 	 pages={28102--28112},
	  year={2025},
	 publisher={IEEE}
}

@article{SAGE,
  	title={Every SAM Drop Counts: Embracing Semantic Priors for Multi-Modality Image Fusion and Beyond},
  	author={Wu, Guanyao and Liu, Haoyu and Fu, Hongming and Peng, Yichuan and Liu, Jinyuan and Fan, Xin and Liu, Risheng},
 	 journal={Proceedings of the IEEE/CVF Conference on Computer Vision and Pattern Recognition},
 	 volume={},
 	 number={},
  	pages={17882--17891},
 	 year={2025},
	 publisher={IEEE}
}

@article{FreeFusion ,
  	title={FreeFusion: Infrared and Visible Image Fusion via Cross Reconstruction Learning}, 
  	author={Zhao, Wenda and Cui, Hengshuai and Wang, Haipeng and He, You and Lu, Huchuan},
  	journal={IEEE Transactions on Pattern Analysis and Machine Intelligence}, 
  	volume={},
 	 number={},
  	pages={1-17},
  	year={2025},
	publisher={IEEE}
}

@article{duochidu1-CM-MCNet,
	title = {CM-MCNet: Convolution and multilayer perceptron-integrated multiscale coordinate network for infrared and visible image fusion},
	author = {Jinping Liu and Shiyi Liu and Lijuan Huang and Lihui Cen},
	journal = {Pattern Recognition},
	volume = {168},
	number={},
	pages = {111843},
	year = {2025},
	publisher={elsevier}
}

@article{duochidu2-MRF-Net,
  	author={Xu, Qizhi and Chen, Jiuchen and Yan, Xinyu and Li, Wei},
  	journal={IEEE Transactions on Geoscience and Remote Sensing}, 
  	title={MRF-Net: An Infrared Remote Sensing Image Thin Cloud Removal Method With the Intra-Inter Coherent Constraint}, 
  	volume={62},
  	number={},
  	pages={1-19},
  	year={2024},
	publisher={IEEE}
}

@article{jin1-WANG2024102414,
	title = {A general image fusion framework using multi-task semi-supervised learning},
	journal = {Information Fusion},
	volume = {108},
  	volume={},
  	number={},
	pages = {102414},
	year = {2024},
	issn = {1566-2535},
	author = {Wu Wang and Liang-Jian Deng and Gemine Vivone},
	publisher={elsevier}
}

@article{jin2-LEITE2025103339,
	title = {Multimodal information fusion using pyramidal attention-based convolutions for underwater tri-dimensional scene reconstruction},
	journal = {Information Fusion},
	volume = {124},
	pages = {103339},
	year = {2025},
	issn = {1566-2535},
  	number={},
	author = {Pedro Nuno Leite and Andry Maykol Pinto},
	publisher={elsevier}
}

@article{
cui2025adair,
title={Ada{IR}: Adaptive All-in-One Image Restoration via Frequency Mining and Modulation},
author={Yuning Cui and Syed Waqas Zamir and Salman Khan and Alois Knoll and Mubarak Shah and Fahad Shahbaz Khan},
booktitle={The Thirteenth International Conference on Learning Representations},
year={2025},
}

@article{AWRaCLe, 
title={AWRaCLe: All-Weather Image Restoration Using Visual In-Context Learning}, 
journal={Proceedings of the AAAI Conference on Artificial Intelligence}, 
author={Rajagopalan, Sudarshan and Patel, Vishal M.}, 
year={2025}, 
month={Apr.}, 
pages={6675-6683},
volume={39},
number={6}
 }

@InProceedings{IVIF3,
    author    = {Geng, Mengyue and Zhu, Lin and Wang, Lizhi and Zhang, Wei and Xiong, Ruiqin and Tian, Yonghong},
    title     = {Event-based Visible and Infrared Fusion via Multi-task Collaboration},
    booktitle = {2024 IEEE/CVF Conference on Computer Vision and Pattern Recognition (CVPR)},
    month     = {June},
    year      = {2024},
    volume = {},
    number={},
    pages     = {26929-26939},
publisher={IEEE}
}

@ARTICLE{IVIF6,
  author={Chen, Jun and Yang, Liling and Yu, Wei and Gong, Wenping and Cai, Zhanchuan and Ma, Jiayi},
  journal={IEEE Transactions on Image Processing}, 
  title={SDSFusion: A Semantic-Aware Infrared and Visible Image Fusion Network for Degraded Scenes}, 
  year={2025},
  volume={34},
  number={},
  pages={3139-3153},
publisher={IEEE}
}

@ARTICLE{IVIF7,
  author={Ma, Haolong and Li, Hui and Cheng, Chunyang and Wang, Gaoang and Song, Xiaoning and Wu, Xiao-Jun},
  journal={IEEE Transactions on Image Processing}, 
  title={S4Fusion: Saliency-aware Selective State Space Model for Infrared and Visible Image Fusion}, 
  year={2025},
  volume={},
  number={},
  pages={1-1},
publisher={IEEE}
}

@article{kungong3,
author = {Huafeng Li and Junyu Liu and Yafei Zhang and Yu Liu},
title = {A Deep Learning Framework for Infrared and Visible Image Fusion
Without Strict Registration},
journal = {International Journal of Computer Vision},
year = {2024},volume={132}, pages={1625–1644},}

@article{kungong-pami,
  title={MulFS-CAP: Multimodal fusion-supervised cross-modality alignment perception for unregistered infrared-visible image fusion},
  author={Li, Huafeng and Yang, Zengyi and Zhang, Yafei and Jia, Wei and Yu, Zhengtao and Liu, Yu},
  journal={IEEE Transactions on Pattern Analysis and Machine Intelligence},
  year={2025},
  publisher={IEEE}
}

@article{xiaosong1,
  title={Multimodal medical image fusion based on joint bilateral filter and local gradient energy},
  author={Li, Xiaosong and Zhou, Fuqiang and Tan, Haishu and Zhang, Wanning and Zhao, Congyang},
  journal={Information Sciences},
  volume={569},
  pages={302--325},
  year={2021},
  publisher={Elsevier}
}

@article{xiaosong2,
  title={Joint image fusion and denoising via three-layer decomposition and sparse representation},
  author={Li, Xiaosong and Zhou, Fuqiang and Tan, Haishu},
  journal={Knowledge-Based Systems},
  volume={224},
  pages={107087},
  year={2021},
  publisher={Elsevier}
}

@inproceedings{TG,
  title={Task-Gated Multi-Expert Collaboration Network for Degraded Multi-Modal Image Fusion},
  author={Sun, Yiming and Li, Xin and Zhu, Pengfei and Hu, Qinghua and Ren, Dongwei and Xu, Huiying and Zhu, Xinzhong},
  booktitle={Forty-second International Conference on Machine Learning}
}

@article{kong4,
title = {Automatic body temperature detection of group-housed piglets based on infrared and visible image fusion},
journal = {Artificial Intelligence in Agriculture},
volume = {16},
number = {1},
pages = {1-11},
year = {2026},
issn = {2589-7217},
author = {Kaixuan Cuan and Feiyue Hu and Xiaoshuai Wang and Xiaojie Yan and Yanchao Wang and Kaiying Wang},
}

@article{kong5,
title = {A three-stage model for infrared small target detection with spatial and semantic feature fusion},
journal = {Expert Systems with Applications},
volume = {295},
pages = {128776},
year = {2026},
issn = {0957-4174},
author = {Sixiang Ji and Haofei Zhang and Jingmin Zhang and Chun Fei and Xiaoyang Wang and Juanxiu Liu and Ping Zhang},
}

@INPROCEEDINGS{M3FD,
  author={Liu, Jinyuan and Fan, Xin and Huang, Zhanbo and Wu, Guanyao and Liu, Risheng and Zhong, Wei and Luo, Zhongxuan},
  booktitle={2022 IEEE/CVF Conference on Computer Vision and Pattern Recognition (CVPR)}, 
  title={Target-aware Dual Adversarial Learning and a Multi-scenario Multi-Modality Benchmark to Fuse Infrared and Visible for Object Detection}, 
  year={2022},
  volume={},
  number={},
  pages={5792-5801},
publisher={IEEE}
}

@article{MSRS,
title = {PIAFusion: A progressive infrared and visible image fusion network based on illumination aware},
journal = {Information Fusion},
volume = {83-84},
pages = {79-92},
year = {2022},
issn = {1566-2535},
author = {Linfeng Tang and Jiteng Yuan and Hao Zhang and Xingyu Jiang and Jiayi Ma},
publisher={elsevier}
}

@INPROCEEDINGS{LLVIP,
  author={Jia, Xinyu and Zhu, Chuang and Li, Minzhen and Tang, Wenqi and Zhou, Wenli},
  booktitle={2021 IEEE/CVF International Conference on Computer Vision Workshops (ICCVW)}, 
  title={LLVIP: A Visible-infrared Paired Dataset for Low-light Vision}, 
  year={2021},
  volume={},
  number={},
  pages={3489-3497},
publisher={IEEE}
}

@article{325-tip,
  title={UMCFuse: A Unified Multiple Complex Scenes Infrared and Visible Image Fusion Framework},
  author={Li, Xilai and Li, Xiaosong and Tan, Tianshu and Li, Huafeng and Ye, Tao},
  journal={IEEE Transactions on Image Processing},
  year={2025},
  publisher={IEEE}
}

@article{325-yixue,
  title={FlexiD-Fuse: Flexible number of inputs multi-modal medical image fusion based on diffusion model},
  author={Xu, Yushen and Li, Xiaosong and Wang, Yuchun and Cheng, Xiaoqi and Li, Huafeng and Tan, Haishu},
  journal={Expert Systems with Applications},
  pages={128895},
  year={2025},
  publisher={Elsevier}
}

@article{325-yixue1,
  title={Generative adversarial network for trimodal medical image fusion using primitive relationship reasoning},
  author={Huang, Jingxue and Li, Xiaosong and Tan, Haishu and Cheng, Xiaoqi},
  journal={IEEE Journal of Biomedical and Health Informatics},
  year={2024},
  publisher={IEEE}
}

@inproceedings{325-xu,
  title={Simultaneous tri-modal medical image fusion and super-resolution using conditional diffusion model},
  author={Xu, Yushen and Li, Xiaosong and Jie, Yuchan and Tan, Haishu},
  booktitle={International Conference on Medical Image Computing and Computer-Assisted Intervention},
  pages={635--645},
  year={2024},
  organization={Springer}
}

@article{325-my,
  title={AWM-Fuse: Multi-Modality Image Fusion for Adverse Weather via Global and Local Text Perception},
  author={Li, Xilai and Liu, Huichun and Li, Xiaosong and Ye, Tao and Kuang, Zhenyu and Li, Huafeng},
  journal={arXiv preprint arXiv:2508.16881},
  year={2025}
}

@article{ESWA-jie,
  title={Medical image fusion based on extended difference-of-Gaussians and edge-preserving},
  author={Jie, Yuchan and Li, Xiaosong and Zhou, Fuqiang and Tan, Haishu and others},
  journal={Expert Systems with Applications},
  volume={227},
  pages={120301},
  year={2023},
  publisher={Elsevier}
}

@article{ESWA-ronghe,
  title={FefDM-Transformer: Dual-channel multi-stage Transformer-based encoding and fusion mode for infrared--visible images},
  author={Li, Junwu and Wang, Yaomin and Ning, Xin and He, Wenguang and Cai, Weiwei},
  journal={Expert Systems with Applications},
  volume={277},
  pages={127229},
  year={2025},
  publisher={Elsevier}
}

@article{ESWA-ronghe1,
  title={MFS-Fusion: Mamba-Integrated Deep Multi-Modal Image Fusion Framework with Multi-Scale Fourier Enhancement and Spatial Calibration},
  author={Chen, Tao and Wang, Chuang and Zhang, Yuanpeng and Xia, Kaijian and Qian, Pengjiang},
  journal={Expert Systems with Applications},
  pages={130054},
  year={2025},
  publisher={Elsevier}
}

@inproceedings{Re-1,
  title={Restormer: Efficient transformer for high-resolution image restoration},
  author={Zamir, Syed Waqas and Arora, Aditya and Khan, Salman and Hayat, Munawar and Khan, Fahad Shahbaz and Yang, Ming-Hsuan},
  booktitle={Proceedings of the IEEE/CVF conference on computer vision and pattern recognition},
  pages={5728--5739},
  year={2022}
}

@article{re-2,
  title={Promptir: Prompting for all-in-one image restoration},
  author={Potlapalli, Vaishnav and Zamir, Syed Waqas and Khan, Salman H and Shahbaz Khan, Fahad},
  journal={Advances in Neural Information Processing Systems},
  volume={36},
  pages={71275--71293},
  year={2023}
}

@inproceedings{re-3,
  title={Mambair: A simple baseline for image restoration with state-space model},
  author={Guo, Hang and Li, Jinmin and Dai, Tao and Ouyang, Zhihao and Ren, Xudong and Xia, Shu-Tao},
  booktitle={European conference on computer vision},
  pages={222--241},
  year={2024},
  organization={Springer}
}

@inproceedings{re-4,
  title={Swinir: Image restoration using swin transformer},
  author={Liang, Jingyun and Cao, Jiezhang and Sun, Guolei and Zhang, Kai and Van Gool, Luc and Timofte, Radu},
  booktitle={Proceedings of the IEEE/CVF international conference on computer vision},
  pages={1833--1844},
  year={2021}
}

@inproceedings{re-5,
  title={Channel Consistency Prior and Self-Reconstruction Strategy Based Unsupervised Image Deraining},
  author={Dong, Guanglu and Zheng, Tianheng and Cao, Yuanzhouhan and Qing, Linbo and Ren, Chao},
  booktitle={Proceedings of the Computer Vision and Pattern Recognition Conference},
  pages={7469--7479},
  year={2025}
}

@inproceedings{re-6,
  title={Interaction-Guided Two-branch image dehazing network},
  author={Liu, Huichun and Li, Xiaosong and Tan, Tianshu},
  booktitle={Proceedings of the Asian Conference on Computer Vision},
  pages={4069--4084},
  year={2024}
}



\end{document}